\documentclass{article}

\PassOptionsToPackage{numbers, compress}{natbib}
\usepackage[preprint]{neurips_2026}

\usepackage[utf8]{inputenc} 
\usepackage[T1]{fontenc}    
\usepackage{hyperref}       
\usepackage{url}            
\usepackage{booktabs}       
\usepackage{amsfonts}       
\usepackage{nicefrac}       
\usepackage{microtype}      
\usepackage[table]{xcolor}         
\usepackage{graphicx}
\usepackage{amsmath}
\usepackage{subcaption}
\usepackage{multirow}
\usepackage{enumitem}
\usepackage{placeins}

\title{ECHO: Continuous Hierarchical Memory for Vision-Language-Action Models}

\author{%
  \textbf{Yanbin Hu}$^{1,3}$,
  \textbf{Jin Cui}$^{2,3}$,
  \textbf{Jiayi Lu}$^{3}$,
  \textbf{Ruixuan Yang}$^{3}$,
  \textbf{Jun Ye}$^{3}$, \\
  \textbf{Boran Zhao}$^{*,1,3}$,
  \textbf{Xingyu Chen}$^{2,3}$,
  \textbf{Xuguang Lan}$^{2,3}$,
  \textbf{Pengju Ren}$^{2,3}$ \\
  \\
  $^{1}$School of Software, Xi'an Jiaotong University \\
  $^{2}$School of Artificial Intelligence, Xi'an Jiaotong University \\
  $^{3}$State Key Laboratory of Human-Machine Hybrid Augmented Intelligence, \\
  Institute of Artificial Intelligence and Robotics, Xi'an Jiaotong University \\
  \texttt{yanbinhu@stu.xjtu.edu.cn} \quad
  \texttt{\{boranzhao, pengjuren\}@xjtu.edu.cn}
}

\begin{document}

\maketitle

\begingroup
\renewcommand{\thefootnote}{*}
\footnotetext{Corresponding author: \texttt{boranzhao@xjtu.edu.cn}.}
\endgroup

\begin{abstract}

Memory capacity is a critical factor determining the performance of Vision-Language-Action (VLA) models in long-horizon manipulation tasks. Existing memory-augmented architectures primarily rely on linear or flat storage, lacking structural priors for manipulation categories and hierarchical organization. This deficiency hinders efficient experience retrieval and limits generalization to unseen long-horizon task compositions. Inspired by the hierarchical organization of human experience, we propose \textbf{\textit{ECHO}} \textit{(\textbf{E}xperience \textbf{C}onsolidation and \textbf{H}ierarchical \textbf{O}rganization)}, a novel memory framework operating within a Continuous Hierarchical Space. By employing a hyperbolic autoencoder, \textit{ECHO} maps VLA hidden states into this space. Leveraging hyperbolic metrics and entailment constraint mechanisms, experience vectors are organized into a semantic memory tree that supports efficient top-down retrieval. In parallel, a background consolidation mechanism continuously refines the memory tree through geometric interpolation and structural splitting, supporting virtual memory synthesis in the continuous space. We integrate \textit{ECHO} into the $\pi_0$ foundation model. Evaluations on LIBERO and preliminary real-world experiments demonstrate the effectiveness of our approach, notably achieving a 12.8\% absolute improvement in execution success rate over the $\pi_0$ baseline on LIBERO-Long, while improving compositional generalization on cross-suite unseen long-horizon tasks.

\end{abstract}
 
\section{Introduction}
\label{sec:introduction}



Driven by rapid advancements in cross-modal environmental understanding, Vision-Language-Action (VLA) models such as RT-2 \cite{rt2}, OpenVLA \cite{openvla}, and $\pi_0$ \cite{pi0} can directly translate visual observations and text instructions into executable robot actions. During this process, integrating a robust memory system into these foundation models is crucial for achieving advanced embodied intelligence\cite{memoryvla,mem,map-vla}. Particularly in complex, long-horizon manipulation tasks, the model's ability to extract and utilize past successful experiences often determines the upper bound of decision-making\cite{dejavu,goal2skill,optimus}. A key property of such experience reuse is hierarchical generalization: when humans encounter a new task, they can often transfer memories from similar categories by abstracting common subgoals and action patterns, rather than replaying a previously seen trajectory. This suggests that useful robotic memory should not only store past episodes, but also organize them into reusable abstractions that support generalization across related manipulation categories. Current mainstream memory-augmented architectures typically employ linear sequences to store historical experiences as unordered temporal slices. However, we observe that this paradigm has structural limitations\cite{vla-survey}. It lacks a structural prior for manipulation categories, failing to capture the inherent containment and subordination relationships in complex tasks. Consequently, it cannot efficiently extract common knowledge from past experiences to guide new, similar tasks.

\begin{figure}[htbp]
  \centering
  \begin{subfigure}[b]{0.48\textwidth}
    \centering
    \includegraphics[width=\textwidth]{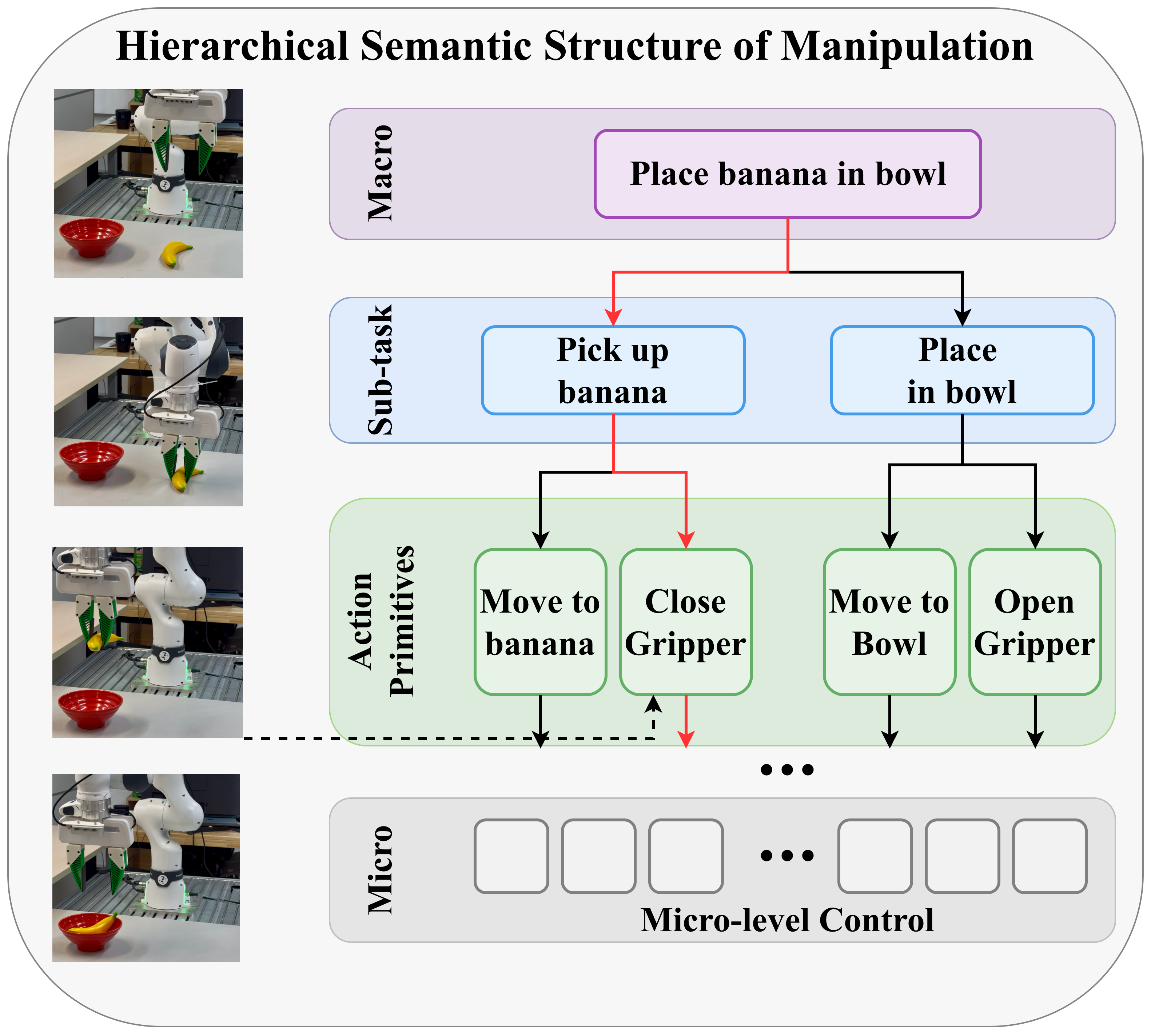}
    \caption{Hierarchical semantics of manipulation tasks}
    \label{fig:hier-semantics}
  \end{subfigure}
  \hfill
  \begin{subfigure}[b]{0.485\textwidth}
    \centering
    \includegraphics[width=\textwidth]{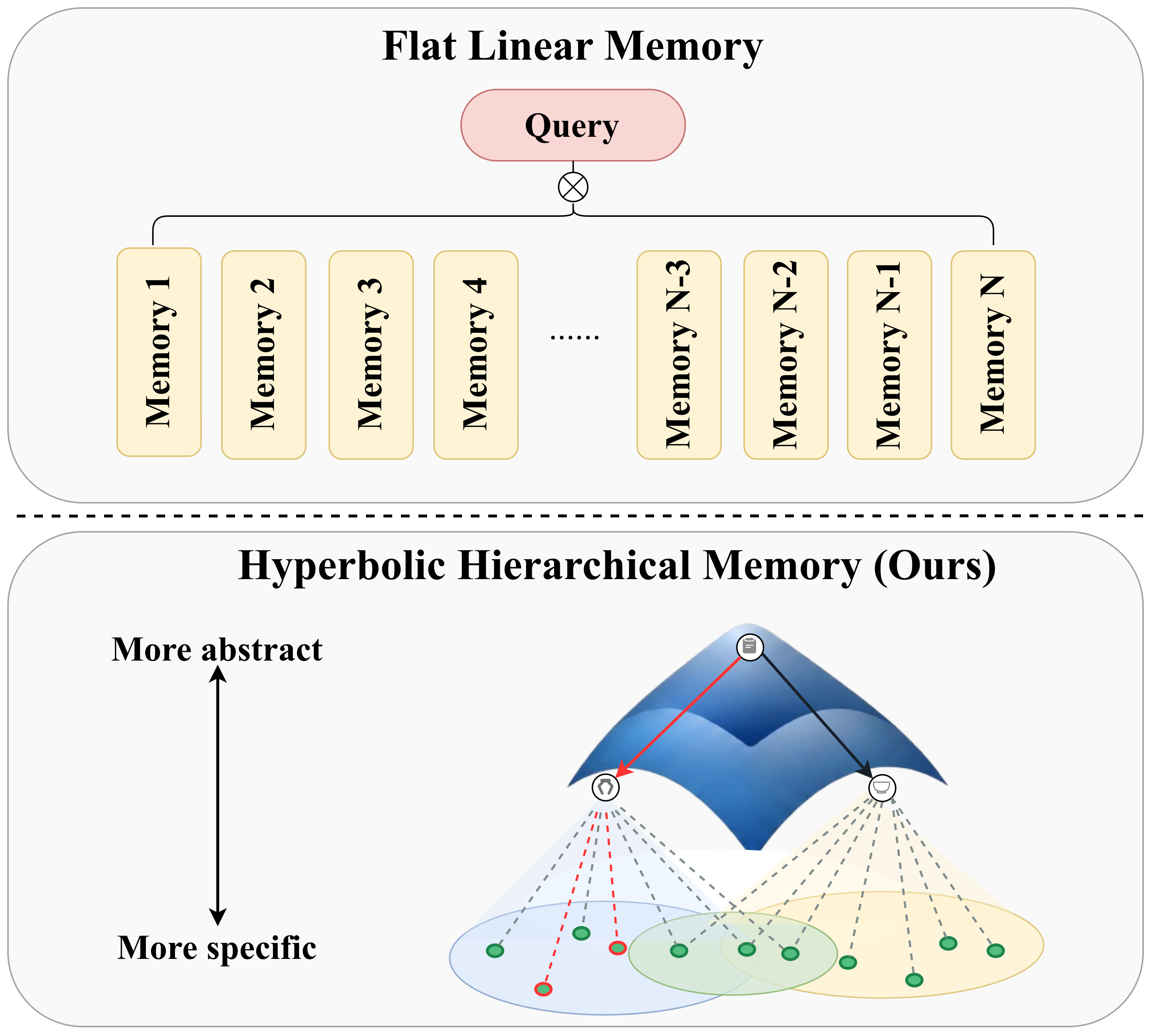}
    \caption{Flat memory vs. hyperbolic hierarchy}
    \label{fig:flat-vs-hyperbolic}
  \end{subfigure}
  \caption{
  Conceptual overview of the \textit{ECHO} framework.
  (a) Manipulation demonstrations exhibit hierarchical semantics, where high-level task instructions are grounded through sub-tasks, action primitives, and micro-level controls.
  (b) Instead of organizing experiences as a flat linear memory, \textit{ECHO} embeds them into a continuous hyperbolic hierarchical space, enabling coarse-to-fine memory organization, entailment-aware retrieval, and the generation of unseen memories through spatial interpolation.
  }
  \label{fig:linear-vs-hie}
\end{figure}

Unlike simple linear sequence storage, human experience is organized through hierarchical semantic clustering\cite{guo2026neural,chen2025human,kumaran2016learning,mcclelland1995there}. Robot manipulation actions naturally possess a tree-like containment relationship ranging from macroscopic semantics to microscopic control, as illustrated in Figure~\ref{fig:linear-vs-hie}. This hierarchical nature motivates us to move beyond conventional flat memory organization and design a long-term memory mechanism that explicitly models the relationships among tasks, sub-goals, and actions.

However, introducing hierarchical memory into VLA models remains challenging. Discrete tree structures easily suffer from combinatorial explosion under fine-grained action variations\cite{ahn2022can}, while continuous spaces require an appropriate geometry to measure distances across different abstraction levels. To address this, we propose a Continuous Hierarchical Space and introduce hyperbolic metrics to organize high-level task semantics and low-level action experiences. Moreover, task relationships in continuous space are not rigid parent-child links, but often partial or overlapping containment relations; for example, placing a banana into a bowl involves both picking and placing. We therefore introduce Hyperbolic Entailment Cones to describe such complex relationships. Finally, the continuous space enables unseen memories to be synthesized through geometric interpolation, providing experiential priors for new long-horizon task compositions. The main contributions of this paper are summarized as follows:


\begin{itemize}[leftmargin=*]
    \item \textbf{Continuous hierarchical memory for VLA models.}
    We propose \textit{ECHO}, a long-term memory framework that organizes manipulation experiences in a Continuous Hierarchical Space. Unlike rigid discrete memory trees, \textit{ECHO} represents task semantics, sub-goals, and action experiences in a shared continuous hierarchy, mitigating combinatorial growth under fine-grained action variations.

    \item \textbf{Hyperbolic organization with entailment-aware retrieval.}
    We instantiate the continuous hierarchy with hyperbolic geometry and introduce entailment-cone constraints to model abstraction levels and partial containment relations among manipulation experiences. This enables coarse-to-fine top-down retrieval, avoiding global matching over a flat memory bank.

    \item \textbf{Compositional memory reuse for long-horizon manipulation.}
    The continuous memory space allows \textit{ECHO} to reuse and interpolate between related experiences, enabling the policy to transfer category-level subgoal and action priors to unseen long-horizon task compositions. Experiments on LIBERO, cross-suite generalization, ablations, and preliminary real-world manipulation show that \textit{ECHO} improves long-horizon execution while maintaining stable policy behavior.
\end{itemize}

\begin{figure}[t]
  \centering
  \includegraphics[width=\linewidth]{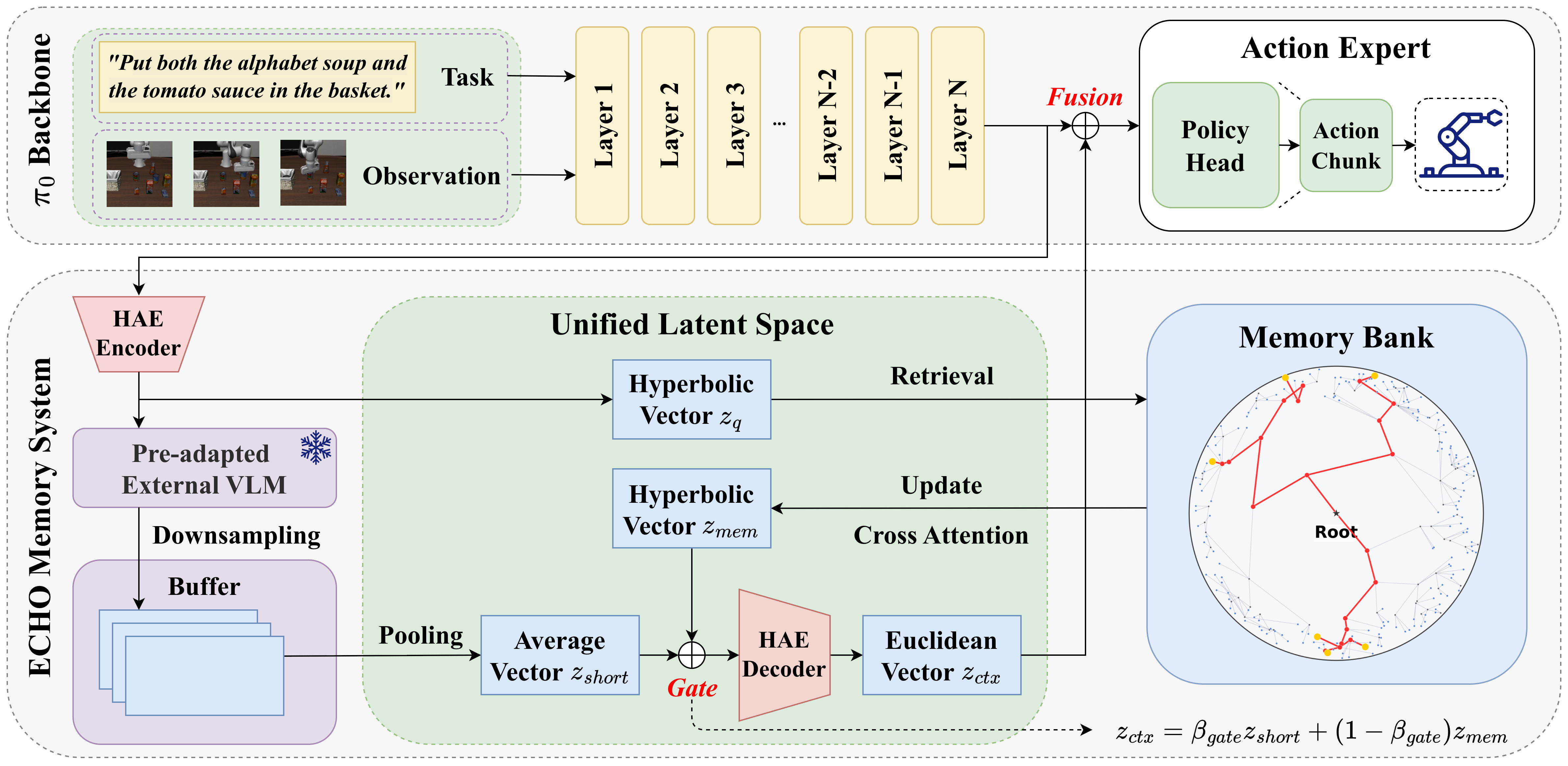}
  \caption{Overall architecture of the \textit{ECHO} framework. The system integrates a hierarchical memory module with the $\pi_0$ backbone. Visual observations are processed by a pre-adapted external VLM, frozen during \textit{ECHO} training, and mapped into a unified hyperbolic latent space. A gating mechanism fuses the short-term context ($z_{\text{short}}$) with the retrieved long-term memory ($z_{\text{mem}}$). The decoded Euclidean context vector ($z_{\text{ctx}}$) is then combined with backbone features to generate action chunks via the action expert.}
  \label{fig:framework}
\end{figure}

\section{Related Work}
\label{sec:related_work}

\paragraph{Vision-Language-Action Models.}
Vision-Language-Action (VLA) models extend Vision-Language Models (VLMs) with action generation capabilities. Existing policies map multimodal inputs to executable actions through diffusion-based decoders \cite{pi0,diffusion-policy,octo}, auto-regressive action prediction \cite{rt2,palm-e}, or continuous action regression \cite{openvla}. While these models have achieved strong reactive control, they still mainly rely on the current observation and local context. As a result, they lack an explicit mechanism for structurally recalling and reusing long-term experiences, which is critical for long-horizon manipulation tasks that require progress tracking across multiple sub-goals.

\paragraph{Memory Systems in Embodied Agents.}
Memory mechanisms have been introduced into embodied agents to improve history awareness and long-horizon decision-making. Some methods maintain short-term history by stacking recent observations within the Transformer context window \cite{rt1, chen2021decision}, while retrieval-augmented approaches retrieve relevant past experiences or expert trajectories from external databases \cite{raea, embodied-rag, statler}. However, these memories are commonly stored as linear sequences or flat vector banks. As interaction data grows, global similarity search over a flat memory bank becomes increasingly expensive \cite{wu2022memorizing}. More importantly, flat retrieval ignores the hierarchical structure of manipulation, where high-level tasks naturally decompose into sub-goals and low-level control primitives \cite{sutton1999between, pertsch2021accelerating}. \textit{ECHO} addresses this limitation by organizing manipulation experiences as a continuous hierarchy rather than merely increasing memory capacity.

\paragraph{Hyperbolic Embeddings for Hierarchical Representation.}
Hyperbolic geometry provides a natural inductive bias for hierarchical representation because negatively curved spaces can embed tree-like structures with low distortion \cite{rthe}. This property has been widely used to model lexical taxonomies such as WordNet \cite{miller1995wordnet,poincare-emb}, graph-structured data \cite{hgcn,hgan}, and continuous hierarchical latent representations \cite{continuous,hyperspherical}. \textit{ECHO} brings this geometric prior to VLA memory organization. By mapping task-conditioned hidden states into a hyperbolic space, \textit{ECHO} represents task-level semantics, sub-goals, and action segments in a unified continuous hierarchy, enabling entailment-aware retrieval and structured experience reuse for robotic manipulation.
\section{Methodology}
\label{sec:method}

\textit{ECHO} is a plug-and-play memory module for VLA policies. Given task-conditioned hidden states from the backbone, \textit{ECHO} compresses successful subgoal segments into a hyperbolic memory space, organizes them as a continuously refined hierarchy, and retrieves relevant experience priors during inference. The retrieved prior is decoded back to the Euclidean feature space and injected into the \(\pi_0\) action pathway through a similarity-modulated residual connection, preserving the stability of the base policy. The overall architecture is shown in Figure~\ref{fig:framework}.

\subsection{Hyperbolic Autoencoder and Entailment Constraints}
\label{subsec:hae}

We use a Hyperbolic Autoencoder (HAE) to map Euclidean latent states from the VLA backbone into the hyperbolic memory space. We adopt the Lorentz model because it provides stable closed-form operations for distance computation, exponential and logarithmic maps, and hierarchy-aware constraints. The Lorentz manifold is defined as $\mathcal{L}^n=\{x\in\mathbb{R}^{n+1}:\langle x,x\rangle_{\mathcal{L}}=-1/c,x_0>0\}$, with the bilinear form:
\begin{equation}
    \langle x, y\rangle_{\mathcal{L}} = -x_0y_0 + \sum_{i=1}^{n} x_i y_i .
\end{equation}

\paragraph{Hierarchy Definition and Entailment Constraints.}
The hierarchical relationship is weakly supervised by the subsumption between global task instructions and sub-goal states. We designate global task embeddings as parent nodes $z_p$ and sub-goal transition states as child nodes $z_c$. To impose this topological prior, the HAE training objective is formulated as:
\begin{equation}
\mathcal{L}_{AE} = \mathcal{L}_{recon}(h, \hat{h}) + \lambda d_{\mathcal{L}}(z_p, z_c) + \gamma \mathcal{L}_{entail},
\end{equation}
where $d_{\mathcal{L}}$ is the Lorentzian distance. The entailment penalty constrains child states to lie inside the entailment cone of their parent task embedding. Specifically, $\phi(z_p,z_c)$ denotes the exterior angle from the parent embedding $z_p$ to the child embedding $z_c$, and $\omega(z_p)$ is the cone half-angle of the parent node. The penalty is activated when the child representation falls outside the parent's entailment cone:
\begin{equation}
\mathcal{L}_{\mathrm{entail}}
=
\frac{1}{|\mathcal{B}|}
\sum_{(z_p,z_c)\in\mathcal{B}}
\max\left(0, \phi(z_p,z_c)-\omega(z_p)\right).
\end{equation}

\paragraph{Parent-child Memory Construction.}
The parent-child pairs do not require manually annotated skill trees. Instead, they are constructed from the task structure. The global language instruction provides the parent-level semantic representation, while sub-goal transition states are extracted by the memory extraction module described in Sec.~\ref{subsec:integration} and used as child-level representations. Each memory entry is represented as a tuple $M_i=(z_i,g_i,s_i,a_i)$, where $z_i \in \mathbb{L}^{n}$ is the hyperbolic embedding, $g_i$ is the global instruction, $s_i$ is the sub-goal description, and $a_i$ denotes the corresponding action chunk or trajectory segment. This design allows \textit{ECHO} to build hierarchical memories from successful rollouts without additional human hierarchy annotations.

\paragraph{Compression and Semantic Recovery.}
After training, the encoder stores memory entries directly as Lorentz vectors, while the decoder maps retrieved memories back to the Euclidean feature space used by the action head. This allows \textit{ECHO} to reduce storage cost while keeping the retrieved memory compatible with the original policy representation.

\subsection{Autonomous Memory Consolidation}
\label{subsec:consolidation}

\textit{ECHO} asynchronously refines the long-term memory tree during robot idle periods. During online execution, newly observed sub-goal transition states are first appended to an unconsolidated memory pool. Once the trajectory passes the task-level consistency check, these experiences are committed to the long-term tree. In this way, successful execution experiences can continuously enrich the memory bank, forming a dynamic experience-learning process without performing expensive structural updates during real-time control.

The structural evolution of the memory tree is governed by two complementary mechanisms: topological expansion and semantic compaction. First, to accommodate novel experiences, new trajectory features $z_{new}$ descend top-down through the hierarchy. A new semantic branch is instantiated if the exterior angle $\phi$ fails the entailment criterion for all existing paths:
\begin{equation}
\forall \mu \in \mathcal{T}_{\mathbb{L}}, \quad \phi(z_{new}, z_\mu) > \omega(z_\mu).
\end{equation}

Conversely, as experiences accumulate within an existing branch, a node $\mu$ may exhibit high internal heterogeneity. If a sufficient fraction of its child entries violates the entailment cone, the node is split:
\begin{equation}
r_{\mathrm{viol}}(\mu)
=
\frac{1}{|Z_\mu|}
\sum_{z\in Z_\mu}
\mathbb{I}\!\left[\phi(z_\mu,z)>\omega(z_\mu)\right]
>
\tau_{\mathrm{split}} .
\end{equation}
Specifically, the system performs a Lorentzian K-Means split to progressively deepen the hierarchy and refine semantic granularity. This process allows overly broad macro-action memories to be decomposed into more precise micro-action memories.

Specifically, the system performs a Lorentzian K-Means split\cite{ganea2018hyperbolic} to progressively deepen the hierarchy and refine semantic granularity. This process allows overly broad macro-action memories to be decomposed into more precise micro-action memories.

\paragraph{Geometric Memory Synthesis.}
The continuous hyperbolic space also supports compositional memory reuse via geometric interpolation. When the current task partially matches multiple stored experiences but has no exact memory counterpart, \textit{ECHO} constructs a temporary virtual memory along the Lorentzian geodesic between relevant memories. For two retrieved memories $z_i$ and $z_j$, the synthesized representation is computed as:
\begin{equation}
z_{syn} = \mathrm{Exp}_{z_i}\left(\rho \, \mathrm{Log}_{z_i}(z_j)\right),
\end{equation}
where $\rho \in [0,1]$ is determined by the normalized similarity between the query and the retrieved memories. The synthesized memory participates in inference as a temporary action prior and is not directly inserted into the memory tree. Only if the resulting execution trajectory passes the consistency check will the corresponding real experience be committed to the long-term memory bank.

\subsection{Hierarchical Retrieval and Alignment}
\label{subsec:retrieval}

\paragraph{Hyperbolic Indexing and Hierarchical Beam Search.}
The retrieval process begins with a top-down search in the long-term memory tree. Given a query vector $q \in \mathbb{L}^n$ generated from the current task-conditioned latent state, the system employs hierarchical beam search to recursively traverse paths satisfying the geometric entailment constraint. Whether a candidate node is expanded is determined by the containment relationship between the query and the centroid of the candidate node $z_\mu$, formulated as $\phi(q, z_\mu) \le \omega(z_\mu)$. Through this geometric filtering mechanism, \textit{ECHO} avoids a global linear scan over the entire memory bank and restricts retrieval to a small number of relevant subtrees under a fixed beam width.

\paragraph{Memory alignment.}
The retrieved memory nodes are mapped to the tangent space and aligned with the current latent state through a lightweight cross-attention module. This produces \(z_{\mathrm{mem}}\), a task-conditioned long-term memory prior. Since alignment is performed in the HAE latent space, the module remains independent of the action decoder and can be plugged into the \(\pi_0\) backbone with minimal architectural changes.

\subsection{Plug-and-Play System Integration}
\label{subsec:integration}

To extract key states from redundant continuous control streams and reduce storage overhead, we introduce a lightweight VLM-guided semantic downsampling module. At task initiation, the VLM decomposes the macro-instruction into ordered sub-goals. During execution, the VLM judge detects sub-goal transitions and extracts the corresponding memory segments. At task termination, a global consistency check is applied to filter failed trajectories or pseudo-memories. Within the unified representation space, we average-pool the short-term buffer to obtain the local context representation $z_{short}$ and use a dynamic gating network $\beta_{gate}$ to adaptively fuse it with the long-term memory representation $z_{mem}$:
\begin{equation}
z_{short} = \frac{1}{|\mathcal{B}|} \sum_{h \in \mathcal{B}} E_\phi(h),
\end{equation}
\begin{equation}
z_{ctx} = \beta_{gate} z_{short} + (1 - \beta_{gate}) z_{mem}.
\end{equation}

The fused context representation $z_{\mathrm{ctx}}$ is decoded back to the Euclidean action feature space as $v_{\mathrm{prior}}=D_\psi(z_{\mathrm{ctx}})$. To preserve the stability of the base $\pi_0$ policy, \textit{ECHO} injects this prior through a residual pathway:
\begin{equation}
E'_{\mathrm{suffix}}=E_{\mathrm{suffix}}+\mathrm{proj}\left(\alpha_t(v_{\mathrm{prior}}-h_{\mathrm{last}})\right).
\end{equation}
The effective strength $\alpha_t$ is modulated by the similarity between the current state and the retrieved memory, so poorly aligned memories have limited influence on the base policy.
\section{Experiments}
\label{sec:experiments}

Our empirical evaluation is designed to answer the following questions: (1) How does \textit{ECHO} perform compared with established VLA baselines and the base $\pi_0$ model on standard long-horizon manipulation benchmarks? (2) Can \textit{ECHO} solve unseen long-horizon task compositions by reusing source-suite memories without target-suite demonstrations? (3) How much does each component contribute, including short-term memory, flat memory retrieval, hyperbolic metrics, entailment-cone retrieval, and background consolidation? (4) How does retrieval latency scale with memory size, and how does the memory injection strength affect policy stability?

\subsection{Experimental Setup}
\label{subsec:experiment_setup}

\textbf{Environments and Tasks.}
We conduct experiments primarily on the \textit{LIBERO}~\cite{libero} simulation benchmark and further validate \textit{ECHO} on a real-world robotic platform. All models are evaluated on four standard LIBERO suites: \textit{LIBERO-Spatial}, \textit{LIBERO-Object}, \textit{LIBERO-Goal}, and \textit{LIBERO-Long} (the 10-task sequence). We additionally use \textit{LIBERO-Plus}~\cite{libero-plus} as a supplementary evaluation for high-complexity scene understanding. Real-world experiments are conducted on a Franka Emika Panda robot to assess the deployability of \textit{ECHO} in physical manipulation settings.

\textbf{Baselines.}
We compare \textit{ECHO} with three groups of baselines. First, Octo and OpenVLA are included as representative generalist VLA policies on the standard LIBERO suites. Second, MAP-VLA and MemoryVLA are included because they also introduce memory or experience reuse mechanisms for VLA-style robotic manipulation. Third, Vanilla $\pi_0$ serves as the most controlled backbone baseline, since \textit{ECHO} is implemented on top of the same locally reproduced $\pi_0$ policy. Results for Octo, OpenVLA, and MAP-VLA are taken from their reported settings and therefore mainly serve as reference points. In contrast, Vanilla $\pi_0$, MemoryVLA, and \textit{ECHO} are reproduced or evaluated in our local pipeline, making these comparisons more directly controlled. For ablations, all variants share the same $\pi_0$ backbone, memory bank, retrieval budget, fusion module, and residual injection strategy, differing only in the memory organization, retrieval rule, and consolidation setting.


\textbf{Memory Bank Construction.}
Unless otherwise specified, the \textit{ECHO} memory bank is constructed from successful rollouts. To improve sub-goal extraction quality and reduce memory pollution, we use a fine-tuned Qwen2-VL-7B-Instruct~\cite{qwen2vl} as the VLM evaluator for transition detection and trajectory verification, as detailed in Appendix~\ref{subsec:vlm_subgoal}. This evaluator is used only for memory construction and filtering, and does not directly generate robot actions. For standard LIBERO evaluation, memories are collected only from successful trajectories in the corresponding training suite under the same evaluation protocol, and evaluation rollouts are never inserted into the memory bank before testing. For cross-suite generalization, the memory bank is populated exclusively with successful experiences from LIBERO-Spatial, LIBERO-Object, and LIBERO-Goal, without inserting any LIBERO-Long target-suite trajectories.

\subsection{Main Results}
\label{subsec:main_results}

Table~\ref{tab:main_results} reports success rates on standard LIBERO suites and LIBERO-Plus. Since Octo, OpenVLA, and MAP-VLA are taken from their reported settings, they serve as reference points rather than strictly controlled comparisons. The most controlled comparison is between Vanilla \(\pi_0\) and \textit{ECHO}, which share the same reproduced backbone and evaluation pipeline.

\textit{ECHO} consistently improves over Vanilla \(\pi_0\) across all standard suites. The gain is most pronounced on \textit{LIBERO-Long}, where \textit{ECHO} improves success from \(80.7\%\) to \(93.5\%\), a \(12.8\%\) absolute improvement. This suggests that structured memory is particularly beneficial when the policy must reuse experience across multiple subgoals and maintain long-horizon task progress. On LIBERO-Plus, \textit{ECHO} also improves from \(54.2\%\) to \(56.5\%\), indicating that the memory prior remains useful in more complex scenes.

\begin{table}[t]
  \caption{
    Success rates (\%) on commonly used and challenging manipulation benchmarks with standard deviations reported when available.
    Octo, OpenVLA, and MAP-VLA results are taken from reported settings, while MemoryVLA is evaluated in our local pipeline.
    A dash ("-") denotes that the baseline was not evaluated in the corresponding environment.
}
  \label{tab:main_results}
  \centering
  \small
  \setlength{\tabcolsep}{6pt}
  \begin{tabular}{lccccc}
    \toprule
    \multirow{2}{*}{Method} & \multicolumn{4}{c}{Standard LIBERO Suites} & \multirow{2}{*}{LIBERO-Plus} \\
    \cmidrule(r){2-5}
    & Spatial & Object & Goal & Long-10 & \\
    \midrule
    Octo~\cite{octo} & 78.9$_{\pm \text{1.0}}$ & 85.7$_{\pm \text{0.9}}$ & 84.6$_{\pm \text{0.9}}$ & 51.1$_{\pm \text{1.3}}$ & - \\
    OpenVLA~\cite{openvla} & 84.7$_{\pm \text{0.9}}$ & 88.4$_{\pm \text{0.8}}$ & 79.2$_{\pm \text{1.0}}$ & 53.7$_{\pm \text{1.3}}$ & 17.3$_{\pm \text{3.2}}$ \\
    MAP-VLA~\cite{map-vla} & 96.3 & 98.4 & 95.4 & 83.4$_{\pm \text{0.7}}$ & - \\
    MemoryVLA~\cite{memoryvla} & 98.0$_{\pm \text{0.6}}$ & 97.4$_{\pm \text{0.9}}$ & 96.4$_{\pm \text{1.3}}$ & 92.4$_{\pm \text{1.1}}$ & - \\
    Vanilla $\pi_0$~\cite{pi0} & 97.5$_{\pm \text{1.7}}$ & 97.0$_{\pm \text{1.2}}$ & 92.3$_{\pm \text{2.5}}$ & 80.7$_{\pm \text{2.0}}$ & 54.2$_{\pm \text{2.9}}$ \\
    \midrule
    \rowcolor{gray!15} \textbf{\textit{ECHO} (Ours)} & \textbf{98.3}$_{\pm \text{1.0}}$ & \textbf{98.8}$_{\pm \text{0.5}}$ & \textbf{98.6}$_{\pm \text{1.0}}$ & \textbf{93.5}$_{\pm \text{2.6}}$ & \textbf{56.5}$_{\pm \text{2.0}}$ \\
    \bottomrule
  \end{tabular}
\end{table}

\begin{figure}[t]
  \centering
  \includegraphics[width=0.9\linewidth]{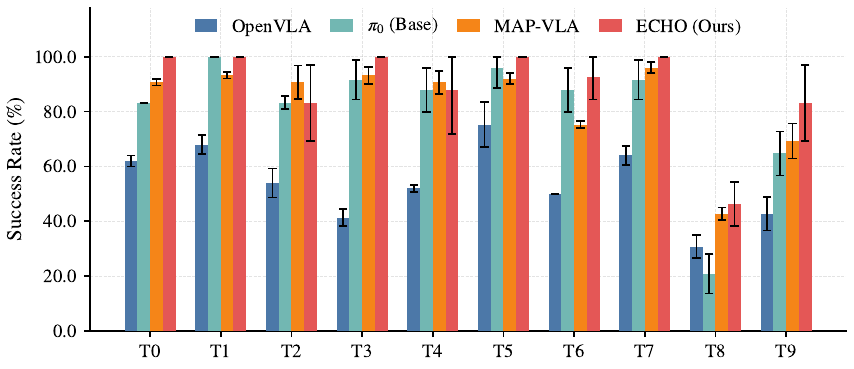}
  \caption{
    Cross-suite compositional generalization on the \textit{LIBERO-Long} 10-task sequence.
    Using only source-suite memories from \textit{LIBERO-Spatial}, \textit{LIBERO-Object}, and \textit{LIBERO-Goal}, \textit{ECHO} improves the average success rate from 80.70\% to 89.31\% without target-suite \textit{LIBERO-Long} memories.
    Error bars denote standard deviations across rollouts and are clipped to $[0,100]$ for visualization.
}
  \label{fig:generalization}
\end{figure}

\subsection{Cross-suite Compositional Generalization}
\label{sec:generalization}

To evaluate cross-suite compositional generalization, we test \textit{ECHO} on the LIBERO-Long 10-task sequence while populating its memory bank exclusively with experiences from \textit{LIBERO-Spatial}, \textit{LIBERO-Object}, and \textit{LIBERO-Goal}. This setting does not include any target-suite LIBERO-Long trajectories, and therefore tests whether \textit{ECHO} can solve unseen long-horizon task compositions by reusing and recombining previously observed atomic skills.

As shown in Figure~\ref{fig:generalization}, under the source-suite memory setting, \textit{ECHO} matches or outperforms Vanilla $\pi_0$ on most tasks. In particular, on bottleneck tasks where the base policy degrades substantially, \textit{ECHO} retrieves relevant action priors and mitigates localized failures. This result indicates that the improvement does not merely come from directly retrieving target-task memories, but from compositional reuse of transferable manipulation experiences.

\begin{figure}[t]
  \centering
  \includegraphics[width=\textwidth]{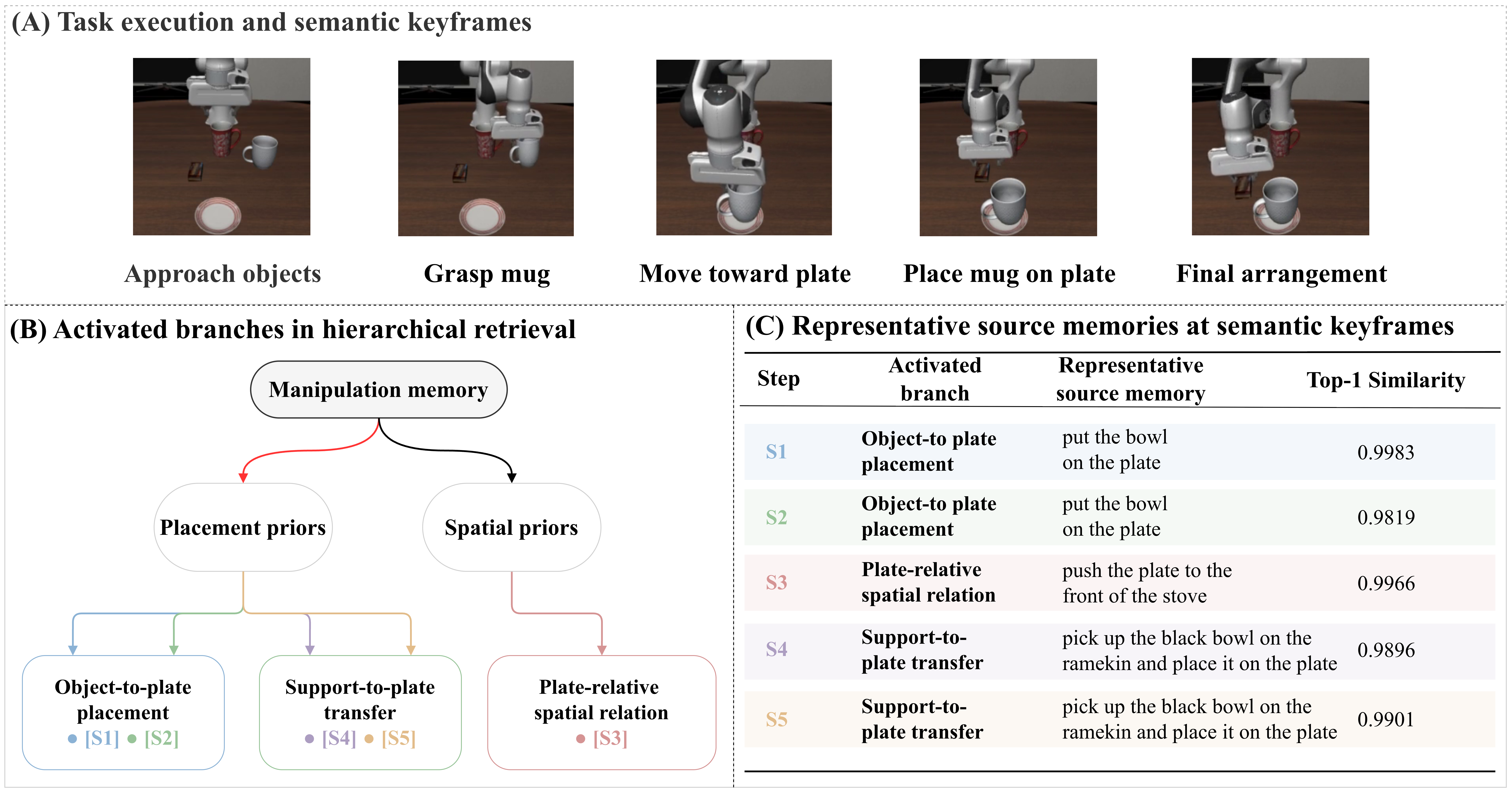}
    \caption{Retrieval-path evidence for cross-suite compositional memory reuse. 
    (A) Semantic keyframes from a successful \textit{ECHO} rollout. 
    (B) Activated branches in the hierarchical memory tree. 
    (C) Representative source-suite memories retrieved at each keyframe. 
    The retrieved entries are not exact target-suite trajectories, but reusable source-task priors such as object-to-plate placement, plate-relative spatial relations, and support-to-plate transfer.}
  \label{fig:qualitative_retrieval}
\end{figure}

\paragraph{Retrieval-path analysis.}
To verify that the improvement comes from compositional reuse rather than direct replay, we visualize a representative successful rollout in Figure~\ref{fig:qualitative_retrieval}. The retrieved memories do not correspond to an exact target-task trajectory; instead, \textit{ECHO} activates reusable source-memory branches, including object-to-plate placement, plate-relative spatial relation, and support-to-plate transfer priors. This suggests that the hierarchical memory tree supports source-suite experience reuse for unseen long-horizon compositions. The complete top-$k$ retrieval records are provided in Appendix~\ref{app:qualitative_retrieval}.

\subsection{Ablation Study}
\label{sec:ablation}

We conduct ablations on \textit{LIBERO-Long} to isolate the contribution of each memory component in \textit{ECHO}. All variants are built on the same reproduced Vanilla $\pi_0$ backbone and share the same memory bank, retrieval budget, fusion module, and residual injection strategy. They differ only in the memory space, retrieval rule, and whether background consolidation is enabled.

The variants isolate the effects of within-episode context, external memory, hyperbolic geometry, entailment-aware tree retrieval, and background consolidation. The short-term-buffer variant uses only episode-level context. The flat-memory variant retrieves the same memory entries with Euclidean kNN. The hyperbolic-memory variant stores memories in the Lorentz manifold but uses distance-only retrieval. Cone Tree Retrieval adds entailment-cone filtering and top-down hierarchical search, while full \textit{ECHO} further enables background consolidation.

\begin{table}[t]
\centering
\caption{Ablation study on \textit{LIBERO-Long}. All memory variants share the same memory bank, retrieval budget, fusion module, and residual injection strategy; they differ only in memory geometry, retrieval rule, and whether consolidation is enabled.}
\label{tab:ablation}
\small
\setlength{\tabcolsep}{4.5pt}
\begin{tabular}{lccc}
\toprule
Model Variant & Memory Space & Retrieval Rule & Success Rate (\%) \\
\midrule
Vanilla $\pi_0$ & -- & -- & 80.70$_{\pm \text{2.01}}$ \\
$\pi_0$ + Short-term Buffer Only & Euclidean & None & 88.81$_{\pm \text{2.01}}$ \\
$\pi_0$ + Flat Memory & Euclidean & kNN & 83.25$_{\pm \text{5.01}}$ \\
$\pi_0$ + Hyperbolic Memory & Hyperbolic & Distance kNN & 91.11$_{\pm \text{3.50}}$ \\
$\pi_0$ + Cone Tree Retrieval & Hyperbolic & Cone Tree & 92.04$_{\pm \text{3.50}}$ \\
\rowcolor{gray!15} \textbf{\textit{ECHO}} & Hyperbolic & Cone Tree + Consolidation & \textbf{93.48}$_{\pm \text{2.89}}$ \\
\bottomrule
\end{tabular}
\end{table}

As shown in Table~\ref{tab:ablation}, the short-term buffer improves the success rate from \(80.70\%\) to \(88.81\%\), indicating that long-horizon manipulation strongly benefits from episode-level state tracking. Flat Euclidean memory improves only modestly to \(83.25\%\) and exhibits larger variance, because flat similarity retrieval may select semantically adjacent but execution-incompatible action priors without hierarchy-aware filtering. In contrast, hyperbolic memory reaches \(91.11\%\), showing that the geometry better organizes reusable experiences across abstraction levels. Cone Tree Retrieval further improves performance to \(92.04\%\), validating the benefit of entailment-cone filtering and top-down hierarchical search. Full \textit{ECHO} achieves the best result of \(93.48\%\), indicating that the final gain comes from the combination of short-term context, hyperbolic organization, structured retrieval, and background consolidation.

\subsection{Real-world Experiments}
\label{subsec:real_world}

To provide preliminary validation of \textit{ECHO} in physical manipulation, we evaluate three representative tabletop manipulation tasks (\textit{Place Banana in Bowl}, \textit{Stack Blocks}, and \textit{Insert Circle into Base}) on a Franka Emika Panda robot. The results show that \textit{ECHO} improves the average real-world success rate from 58.3\% to 70.0\%. Detailed hardware setups, full quantitative result tables, and qualitative rollout sequences are provided in Appendix~\ref{app:real_world_examples}.

\subsection{Memory Organization and System Analysis}
\label{subsec:system_analysis_main}

\begin{figure*}[t]
  \centering

  \begin{subfigure}[t]{0.32\textwidth}
    \centering
    \includegraphics[width=\linewidth]{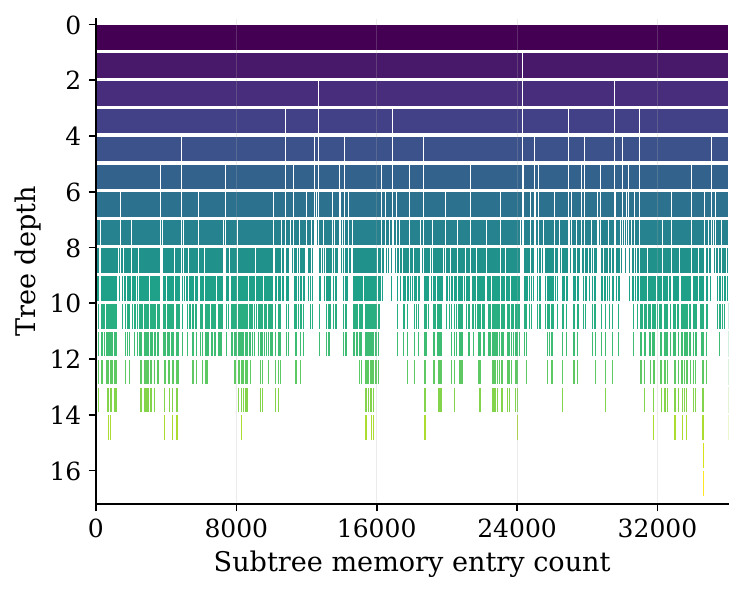}
    \caption{Consolidated memory tree}
    \label{fig:global-memory-tree-a}
  \end{subfigure}
  \hfill
  \begin{subfigure}[t]{0.32\textwidth}
    \centering
    \includegraphics[width=\linewidth]{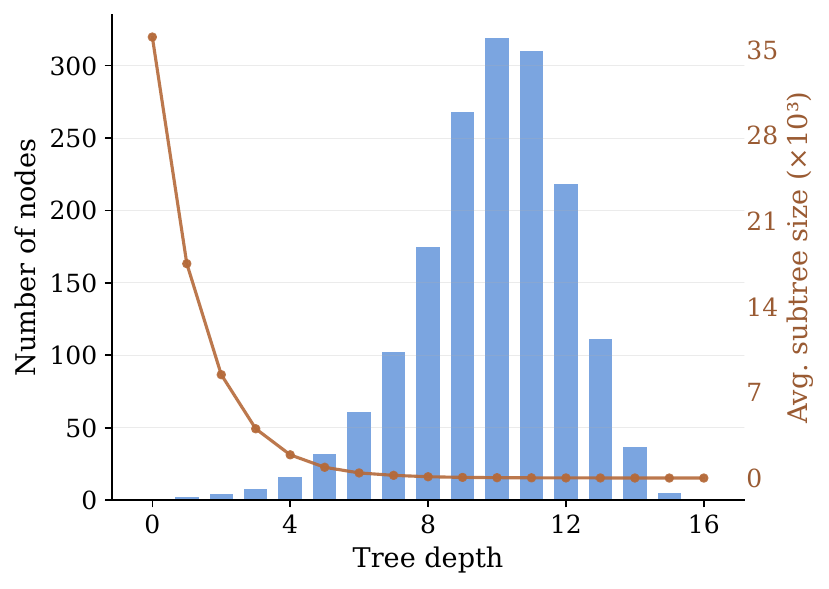}
    \caption{Depth statistics}
    \label{fig:global-memory-tree-b}
  \end{subfigure}
  \hfill
  \begin{subfigure}[t]{0.32\textwidth}
    \centering
    \includegraphics[width=\linewidth]{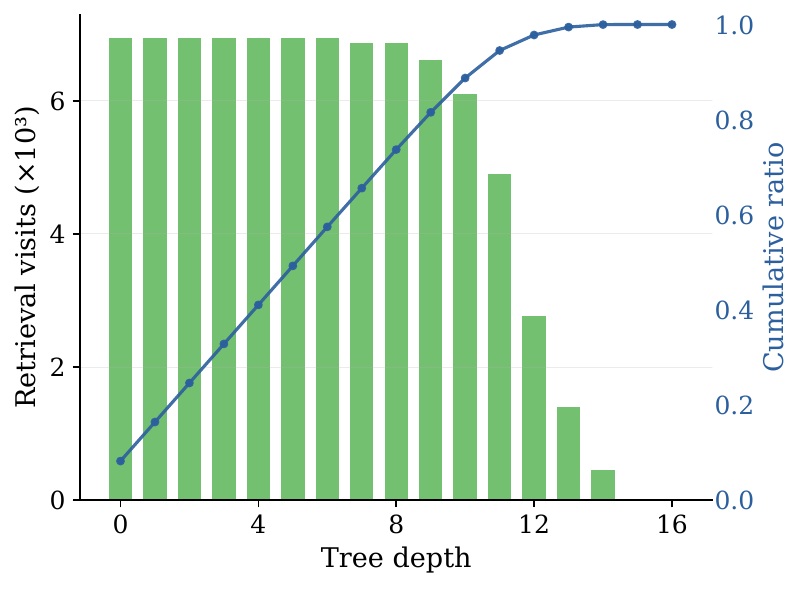}
    \caption{Retrieval visits}
    \label{fig:global-memory-tree-c}
  \end{subfigure}

  \caption{
    Global organization of the consolidated \textit{ECHO} memory bank.
    (a) Icicle visualization of the depth-truncated cone memory tree, where bands denote discrete tree depths and segment widths indicate subtree memory counts.
    (b) Node counts and average subtree sizes across depths, showing how consolidation refines broad clusters into finer semantic branches.
    (c) Retrieval visits over tree depth, showing that \textit{ECHO} performs coarse-to-fine access across multiple hierarchy levels rather than relying on a flat memory pool.
}
\label{fig:memory_tree}
  \label{fig:global-memory-tree}
\end{figure*}

Beyond task success rates, we further analyze the internal memory organization of \textit{ECHO}. Figure~\ref{fig:memory_tree} visualizes the consolidated memory bank. The icicle plot in Figure~\ref{fig:global-memory-tree-a} shows that memory nodes are distributed across multiple depths. Figure~\ref{fig:global-memory-tree-b} indicates that background consolidation refines coarse experience clusters into more fine-grained semantic branches as depth increases. Figure~\ref{fig:global-memory-tree-c} shows that retrieval visits span multiple levels of the hierarchy, suggesting that \textit{ECHO} utilizes the top-down structure for coarse-to-fine access.

Furthermore, we evaluate the impact of base memory injection strength on policy stability and the latency scaling of hyperbolic cone-tree retrieval. The analysis indicates that moderate memory injection maintains policy stability, and cone-tree retrieval exhibits significantly lower latency scaling compared to standard vector search. Detailed quantitative system analysis and related figures are deferred to Appendix~\ref{subsec:system_analysis}.

\section{Conclusion}

We presented \textit{ECHO}, a continuous hierarchical memory framework for VLA models. By organizing successful manipulation experiences in a hyperbolic memory space and retrieving them through entailment-aware tree search, \textit{ECHO} provides reusable long-term priors while preserving the stability of the base \(\pi_0\) policy through residual injection. Experiments on LIBERO, cross-suite generalization, ablations, and preliminary real-world tasks show consistent gains, especially on long-horizon manipulation. Future work will study larger-scale lifelong memory banks, more robust memory verification, and continual deployment in open-ended environments.


\bibliographystyle{unsrtnat}
\bibliography{refs}

\newpage
\appendix
\onecolumn
\section{Core Network Architecture and Hyperparameters}
\label{app:architecture}

\subsection{Hyperbolic Autoencoder Architecture}
The Hyperbolic Autoencoder (HAE) is implemented as a lightweight MLP encoder--decoder. The encoder applies a linear projection, GELU activation, dropout, and a final projection to a 512-dimensional Euclidean latent vector. The decoder uses a symmetric architecture to recover Euclidean action features. To connect Euclidean features with the Lorentz manifold, the encoder applies the exponential map $\exp_0$ as its final operation, while the decoder uses the logarithmic map $\log_0$ before Euclidean reconstruction. Therefore, cached short-term and long-term memory entries are stored directly as hyperbolic representations.

\subsection{Training Details and Hyperparameters}
The HAE is optimized with AdamW using a learning rate of $1\times10^{-3}$, weight decay of $1\times10^{-6}$, and batch size of 256. The hyperbolic latent dimension is $n+1$ with $n=512$, where the additional coordinate corresponds to the time-like dimension of the Lorentz representation. The training objective combines reconstruction loss, Lorentz graph regularization, entailment penalty, and latent norm regularization. The coefficients $\lambda$ and $\gamma$ in the main text correspond to the graph regularization and entailment penalty weights, respectively.

\section{Memory Tree Construction and Asynchronous Consolidation}
\label{app:memory_tree}

\subsection{Lorentzian K-Means Split and Thresholds}
The long-term memory tree is refined with Lorentzian K-Means. We use binary splits ($K=2$) to keep offline consolidation efficient while allowing the hierarchy to become progressively finer as the tree deepens. A node is split when its violation ratio exceeds $\tau_{\mathrm{split}}=0.3$, where the violation ratio is the fraction of child entries whose exterior angle $\phi$ exceeds the cone half-angle $\omega$. This rule prevents overly broad memory nodes from accumulating semantically heterogeneous experiences.

\subsection{Geometric Computation Details}
The cone half-angle $\omega(z)$ is computed from the spatial component of the Lorentz vector $z$ and is clamped near the manifold apex:
\begin{equation}
\omega(z) = \arcsin\left(\frac{2K}{\sqrt{c}\,\|z_{\mathrm{space}}\|_2}\right).
\end{equation}
The exterior angle $\phi$ is computed in the tangent space using the Lorentzian logarithmic map and the Minkowski inner product:
\begin{equation}
\phi = \arccos\left( \frac{\langle u, a \rangle_{\mathcal{L}}}{\|u\|_{\mathcal{L}} \|a\|_{\mathcal{L}}} \right), \quad
u=\log_{z_p}(z_c),\; a=\log_{z_p}(z_{\mathrm{origin}}).
\end{equation}
Here, $z_p$ and $z_c$ denote the parent and child memory representations, respectively.

\subsection{Offline Consolidation Logic}
Background consolidation is executed during robot idle periods. It drains the unconsolidated memory pool, applies cluster splitting when necessary, merges redundant nodes, abstracts repeated semantic patterns, and removes redundant entries. During online inference, new trajectories are only appended to an unconsolidated buffer, avoiding structural tree updates in the real-time control loop.

\section{Hierarchical Retrieval and Feature Alignment}
\label{app:retrieval}

\subsection{Hierarchical Beam Search Parameters}
The retrieval module uses hierarchical beam search with beam width $k=3$. Candidate nodes are ranked by their exterior angle $\phi$, and paths satisfying the entailment constraint $\phi \leq \omega$ are prioritized. This yields sublinear practical retrieval cost under balanced-tree conditions, since only the top-$k$ relevant subtrees are expanded at each depth.

\subsection{Cross-Attention Feature Alignment}
Retrieved hyperbolic memory nodes are first projected to the tangent space through $\log_0$, which removes the time-like Lorentz coordinate and maps each 513-dimensional hyperbolic representation back to a 512-dimensional Euclidean feature. We then apply standard 8-head cross-attention. For batch size $B$ and $K_{\mathrm{ret}}$ retrieved nodes, the query has shape $\mathbb{R}^{B\times512}$ and the tangent-space memory matrix has shape $\mathbb{R}^{B\times(K_{\mathrm{ret}}+1)\times512}$. The per-head dimension is $d_k=64$.

\section{System Integration and Dynamic Network}
\label{app:system_integration}

\subsection{VLM-guided Subgoal Extraction and Verification}
\label{subsec:vlm_subgoal}
We instantiate the VLM evaluator using the open-weight Qwen2-VL-7B-Instruct model. A critical challenge in automated memory construction is "memory pollution"—where incorrect subgoal transitions or suboptimal execution trajectories are erroneously committed to the long-term memory tree. To mitigate this, rather than relying on a zero-shot frozen model, we fine-tune the VLM on a curated dataset of robotic task executions and semantic transition frames.

During inference, the VLM serves a dual purpose: a planner that decomposes long-horizon instructions into visually verifiable subgoals, and a strict judge that detects state transitions. Post-finetuning, the VLM achieves an accuracy of approximately 90\% in semantic transition detection and success verification on seen tasks. All VLM outputs are constrained to structured JSON formats to eliminate parsing ambiguity. This stringent verification process ensures that only high-quality, successful experiences are consolidated into the hyperbolic memory bank.

\subsection{Dynamic Gating and Residual Injection}
The dynamic gating network is a two-layer MLP followed by a sigmoid activation, producing the scalar gate $\beta_{\mathrm{gate}}$ for fusing short-term context and retrieved long-term memory. The residual injection strength is similarity-modulated: the effective coefficient is the product of the base injection strength and a gate derived from the alignment between the current state and the retrieved memory. This prevents poorly matched memories from dominating the base policy.

\section{Simulation Environment and System Configuration}
\label{app:experiments}

\subsection{Simulation Environment Configurations}
For LIBERO evaluation, visual observations are resized to $224\times224$. In LIBERO-Long, tasks are executed according to their official task order to evaluate long-horizon chaining. Unless otherwise specified, the memory bank is built from successful rollouts under the corresponding experimental protocol, and all ablation variants use the same memory entries, retrieval budget, fusion module, and residual injection strategy.


\subsection{System Efficiency and Hyperparameter Analysis}
\label{subsec:system_analysis}

\begin{figure}[h]
  \centering
  \begin{subfigure}{0.48\linewidth}
    \includegraphics[width=\linewidth]{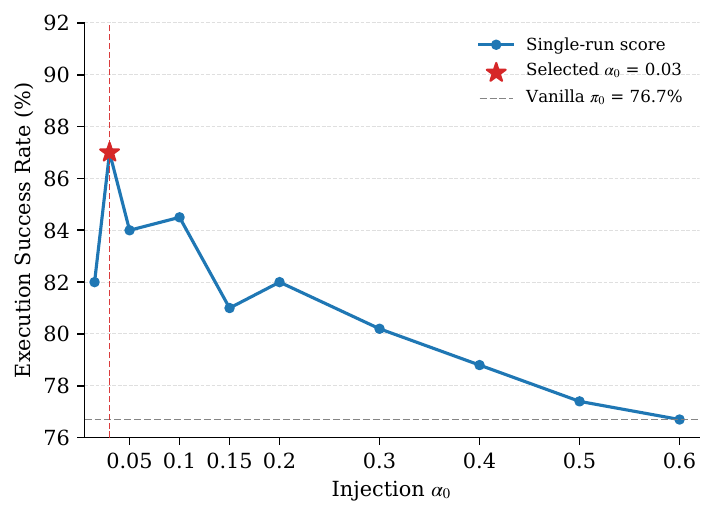}
    \caption{Impact of base injection strength $\alpha_0$.}
    \label{fig:alpha}
  \end{subfigure}
  \hfill
  \begin{subfigure}{0.48\linewidth}
    \includegraphics[width=\linewidth]{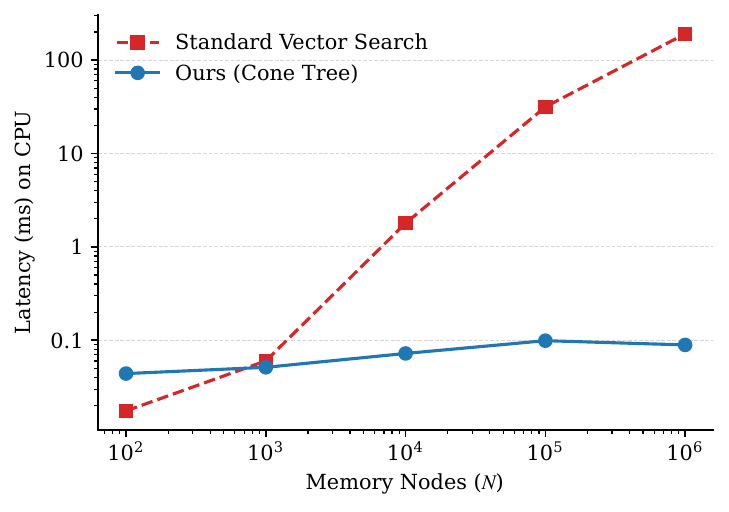}
    \caption{Retrieval latency scaling.}
    \label{fig:latency}
  \end{subfigure}
  \caption{
    System analysis.
    (a) LIBERO-Long success rate as a function of the base memory injection strength $\alpha_0$. The red star indicates the $\alpha_0=0.03$ used for LIBERO-Long experiments. The effective injection strength is further modulated by retrieval-state similarity during execution.
    (b) CPU retrieval latency comparison between standard vector search and hyperbolic cone-tree retrieval as the number of memory nodes ($N$) increases.
  }
  \label{fig:system_analysis}
\end{figure}

This section provides the quantitative analysis of system behavior referenced in the main text. Figure~\ref{fig:system_analysis}(a) analyzes the effect of the base memory injection strength $\alpha_0$ on policy stability. A small $\alpha_0$ limits the influence of memory priors, while an overly large value can make the retrieved memory overly influence the base policy. Therefore, we use a conservative $\alpha_0=0.03$ in the main experiments and further modulate the effective injection strength by retrieval-state similarity.

Figure~\ref{fig:system_analysis}(b) compares CPU retrieval latency between standard vector search and hyperbolic cone-tree retrieval. As the number of memory nodes increases, standard vector search requires matching over a larger global memory pool, whereas cone-tree retrieval expands only relevant subtrees satisfying entailment constraints. This supports the scalability of \textit{ECHO} for larger experience banks.

\subsection{Additional Qualitative Retrieval Case}
\label{app:qualitative_retrieval}

This appendix provides the full top-$k$ retrieval records for the qualitative case study in Figure~\ref{fig:qualitative_retrieval}. The target task is \textit{put the white mug on the plate and put the chocolate pudding to the right of the plate}. \textit{ECHO} succeeds in this episode while the corresponding baseline fails. Retrieved memories are task-level source memories from the retrieval trace; branch labels in the main figure are post-hoc summaries for readability.

\begin{table}[h]
  \centering
  \small
  \setlength{\tabcolsep}{3.5pt}
  \renewcommand{\arraystretch}{1.15}
  \caption{Full top-$k$ retrieval records for the qualitative cross-suite generalization case.}
  \label{tab:qualitative_retrieval_full}
  \resizebox{\textwidth}{!}{
  \begin{tabular}{c p{2.5cm} p{4.2cm} p{4.2cm} p{4.2cm} c}
    \toprule
    Step & Semantic keyframe & Top-1 retrieved memory & Top-2 retrieved memory & Top-3 retrieved memory & Similarity \\
    \midrule
    S1 &
    Approach objects &
    put the bowl on the plate \newline \footnotesize{ID: 2398} &
    put the bowl on the plate \newline \footnotesize{ID: 1956} &
    put the wine bottle on the rack \newline \footnotesize{ID: 2697} &
    1.00 / 1.00 / 1.00 \\

    S2 &
    Grasp mug &
    put the bowl on the plate \newline \footnotesize{ID: 484} &
    put the bowl on the plate \newline \footnotesize{ID: 2443} &
    put the wine bottle on top of the cabinet \newline \footnotesize{ID: 2598} &
    0.98 / 0.98 / 0.97 \\

    S3 &
    Move toward plate &
    push the plate to the front of the stove \newline \footnotesize{ID: 1299} &
    put the wine bottle on top of the cabinet \newline \footnotesize{ID: 997} &
    put the bowl on the stove \newline \footnotesize{ID: 211} &
    1.00 / 1.00 / 1.00 \\

    S4 &
    Place mug on plate &
    pick up the black bowl on the ramekin and place it on the plate \newline \footnotesize{ID: 220} &
    pick up the black bowl on the ramekin and place it on the plate \newline \footnotesize{ID: 572} &
    pick up the black bowl on the ramekin and place it on the plate \newline \footnotesize{ID: 941} &
    0.99 / 0.99 / 0.99 \\

    S5 &
    Final arrangement &
    pick up the black bowl on the ramekin and place it on the plate \newline \footnotesize{ID: 220} &
    pick up the black bowl on the ramekin and place it on the plate \newline \footnotesize{ID: 572} &
    pick up the black bowl on the ramekin and place it on the plate \newline \footnotesize{ID: 941} &
    0.99 / 0.99 / 0.99 \\
    \bottomrule
  \end{tabular}
  }
\end{table}

\FloatBarrier

\subsection{Hardware Specifications}
\label{subsec:hardware}

We report the main hardware and software configurations used in our experiments to support reproducibility. Simulation evaluation and offline memory training were conducted on shared Linux workstations equipped with 8 NVIDIA RTX~6000 Ada GPUs, dual-socket AMD EPYC~7H12 CPUs, and 2.0~TiB host memory. The software environment was built from the project configuration and used PyTorch~2.7.1+cu126, CUDA runtime~12.6, and JAX~0.5.3. LIBERO evaluation uses workspace and wrist camera observations, with visual inputs processed to $224\times224$. The retrieval latency benchmark is implemented with CPU tensors. Real-world experiments are conducted on a Franka Panda-class platform at approximately 15~Hz, with inference served by a remote policy server.

\begin{table}[h]
\centering
\small
\caption{Hardware and software specifications.}
\label{tab:hardware_specs}
\begin{tabular}{p{0.28\linewidth}p{0.63\linewidth}}
\hline
\textbf{Quantity} & \textbf{Specification} \\
\hline
Simulation accelerator(s) & 8 NVIDIA RTX 6000 Ada Generation GPUs, each with 49,140 MiB VRAM \\
Host CPU & AMD EPYC 7H12 64-Core Processor; 256 logical cores \\
Host RAM & 2.0 TiB \\
Software stack & PyTorch 2.7.1+cu126; CUDA runtime 12.6; NVIDIA driver-reported CUDA 13.0 \\
Memory preprocessing & Memory-bank construction, including graph preload, takes approximately 4.1--4.6 minutes per run on the same workstation \\
Physical deployment & Franka Emika Panda class arm; 15 Hz control loop; $224\times224$ RGB inputs with padding and resizing; remote policy-server inference \\
Retrieval latency benchmark & CPU tensor implementation in PyTorch, measured on the AMD EPYC 7H12 host \\
\hline
\end{tabular}
\end{table}

\section{Real-world Setup and Rollout Examples}
\label{app:real_world_examples}

This appendix supplements the real-world experiments discussed in Section~\ref{subsec:real_world} with full setup details, quantitative tables, and qualitative rollout examples.

\textbf{Hardware Platform and Deployment Parameters.} 
Real-world evaluations use a Franka Emika Panda robotic arm with a parallel gripper and a static third-person RGB camera. The control loop targets 15 Hz. RGB observations are padded and resized to $224\times224$ before policy inference, preserving the aspect ratio and avoiding cropping artifacts. The policy server runs remotely through a websocket interface, while proprioceptive states and gripper signals are streamed to match the simulation-side policy input format.

\textbf{Task Evaluation Setup.}
We select three representative tabletop manipulation tasks: \textit{Place Banana in Bowl}, \textit{Stack Blocks}, and \textit{Insert Circle into Base}, covering semantic goal grounding, long-horizon sequential planning, and high-precision spatial control. We evaluate each task with 20 independent trials under randomized initial object configurations.

\begin{figure}[htbp]
  \centering
  \begin{minipage}[c]{0.54\textwidth}
    \centering
    \captionof{table}{Success rates (\%) in real-world manipulation tasks.}
    \label{tab:real_world_results}
    \vspace{0.5em}
    \renewcommand{\arraystretch}{1.4}
    \small
    \setlength{\tabcolsep}{5pt}
    \begin{tabular}{lcccc}
      \toprule
      Model & Place & Stack & Insert & Avg. \\
      \midrule
      Vanilla $\pi_0$ & 75.0 & 45.0 & 55.0 & 58.3 \\
      \rowcolor{gray!15} \textbf{\textit{ECHO} (Ours)} & \textbf{90.0} & \textbf{55.0} & \textbf{65.0} & \textbf{70.0} \\
      \bottomrule
    \end{tabular}
  \end{minipage}
  \hfill
  \begin{minipage}[c]{0.42\textwidth}
    \centering
    \includegraphics[width=\linewidth, trim=0 3cm 0 2cm, clip]{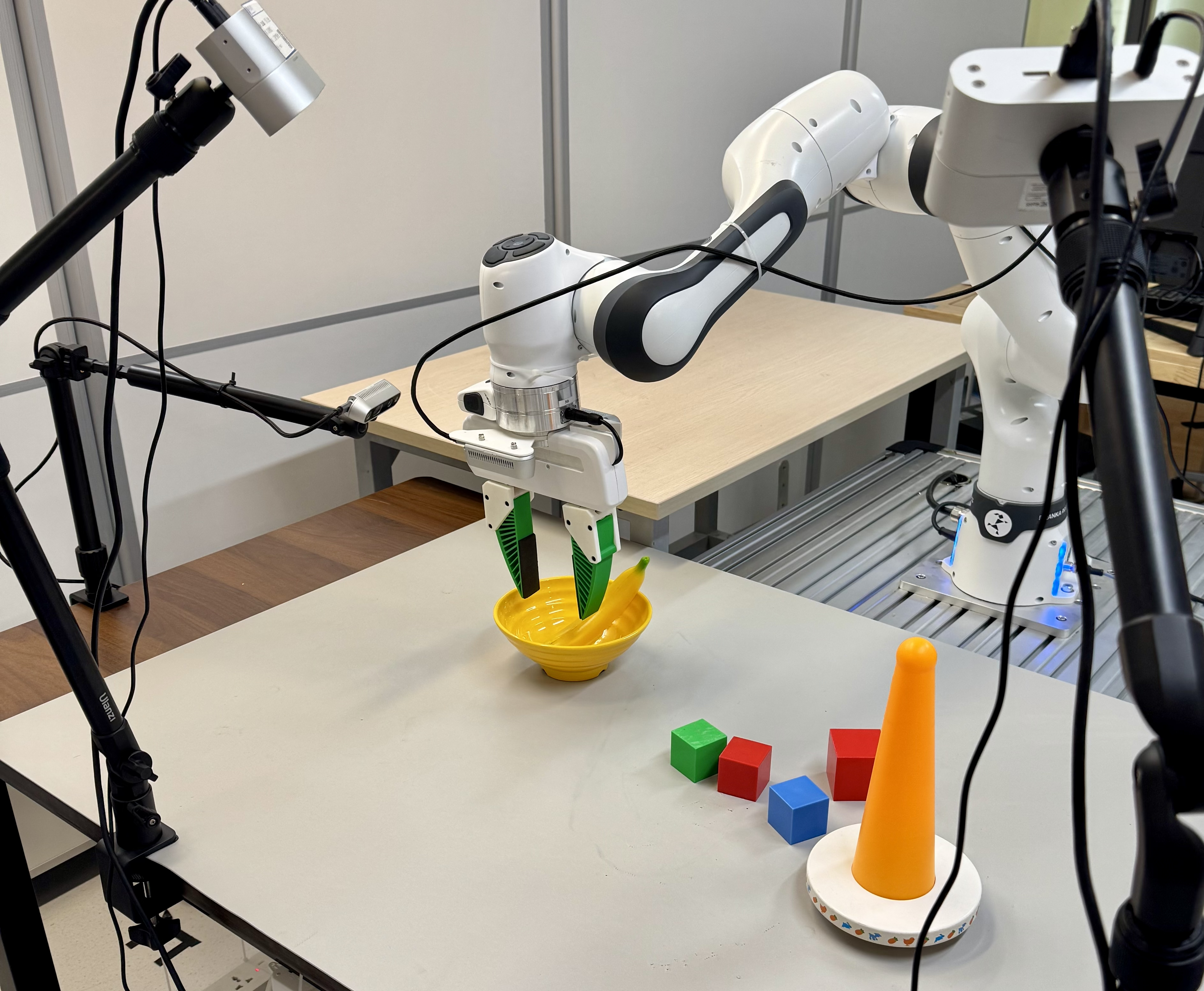}
    \caption{Real-world experimental setup.}
    \label{fig:robot_photo}
  \end{minipage}
\end{figure}

\begin{figure}[htbp]
    \centering
    \begin{subfigure}{0.24\textwidth}
        \includegraphics[width=\linewidth, trim=0 18cm 0 18cm, clip]{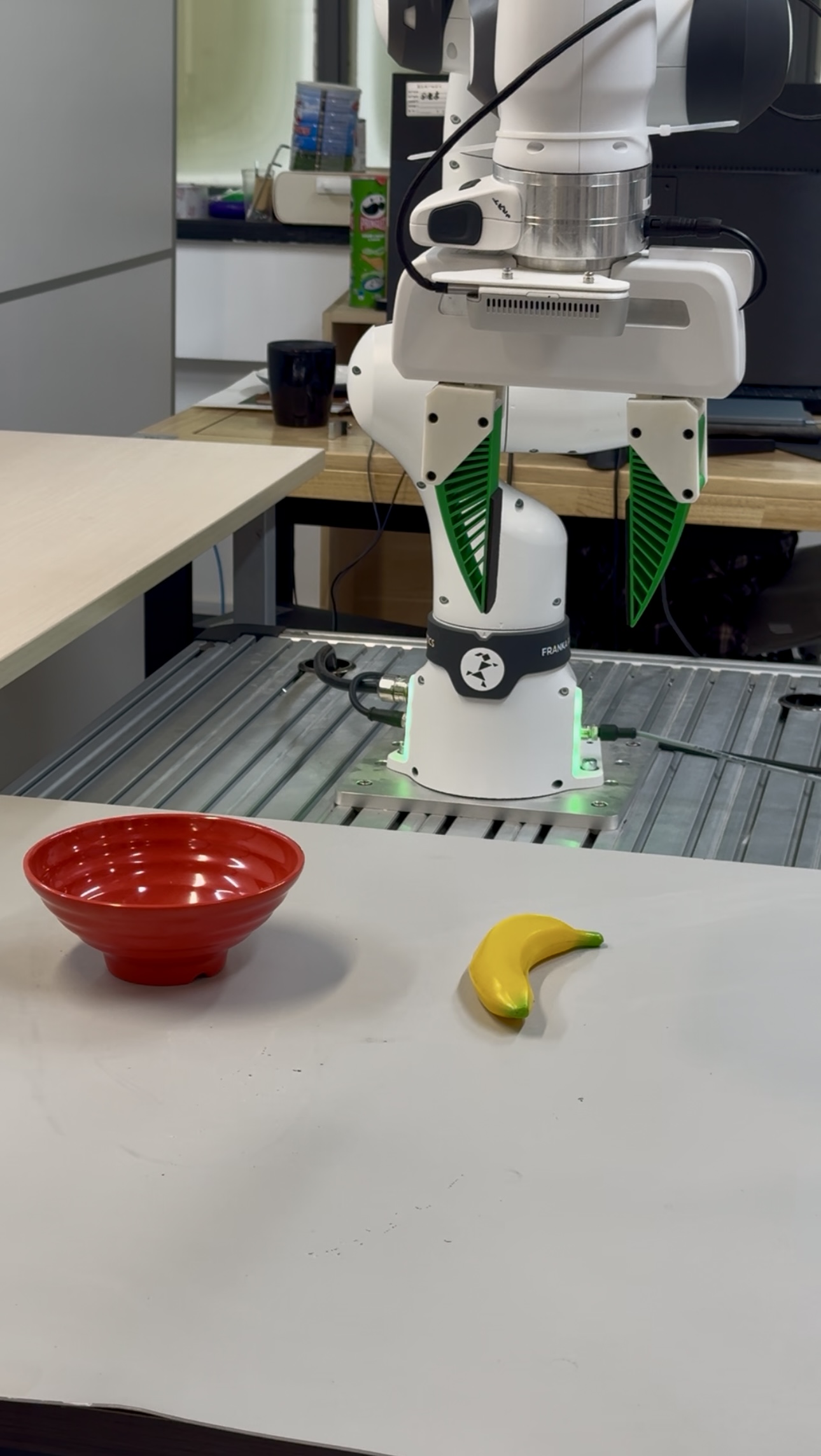}
    \end{subfigure}\hfill
    \begin{subfigure}{0.24\textwidth}
        \includegraphics[width=\linewidth, trim=0 18cm 0 18cm, clip]{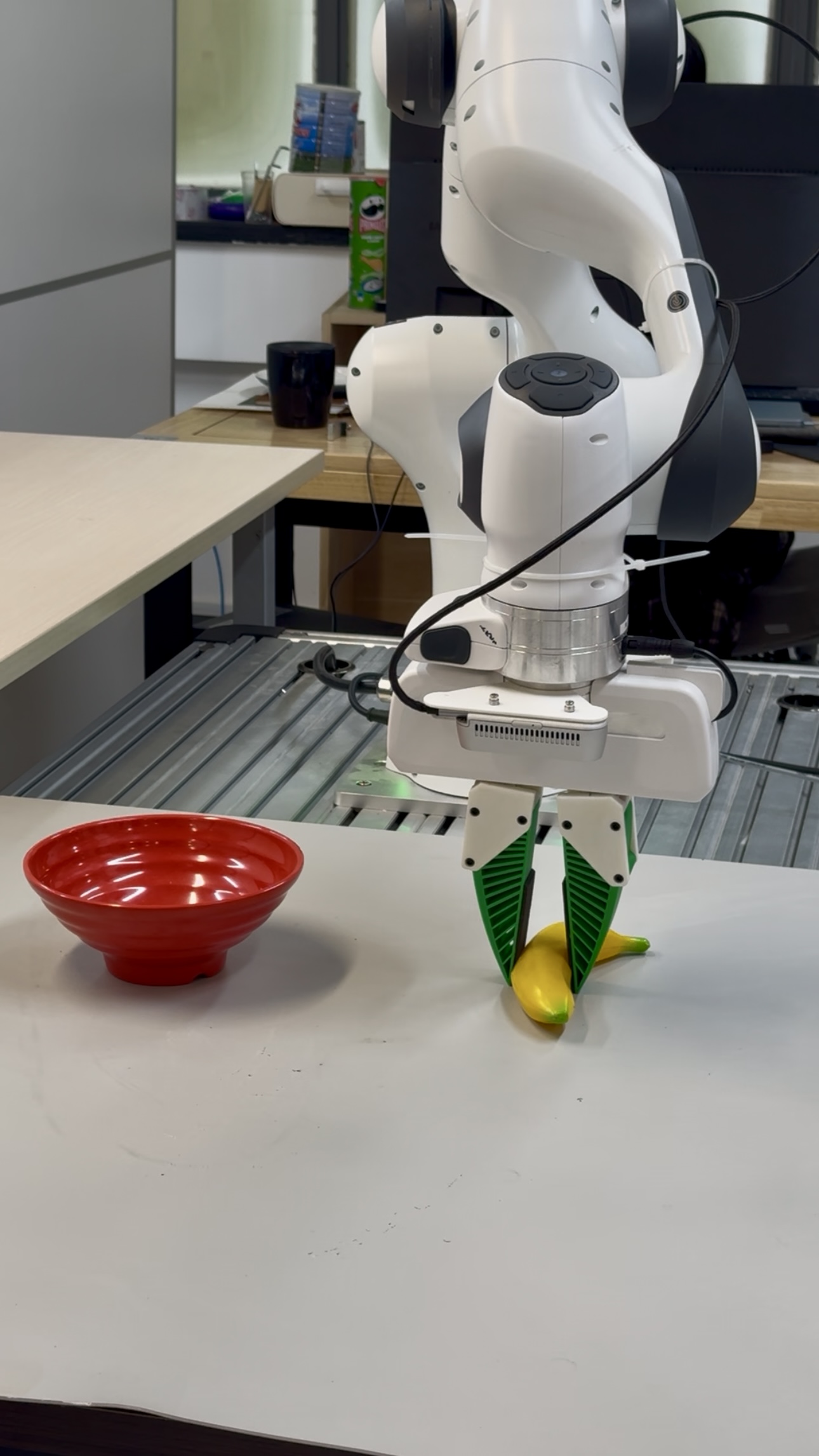}
    \end{subfigure}\hfill
    \begin{subfigure}{0.24\textwidth}
        \includegraphics[width=\linewidth, trim=0 18cm 0 18cm, clip]{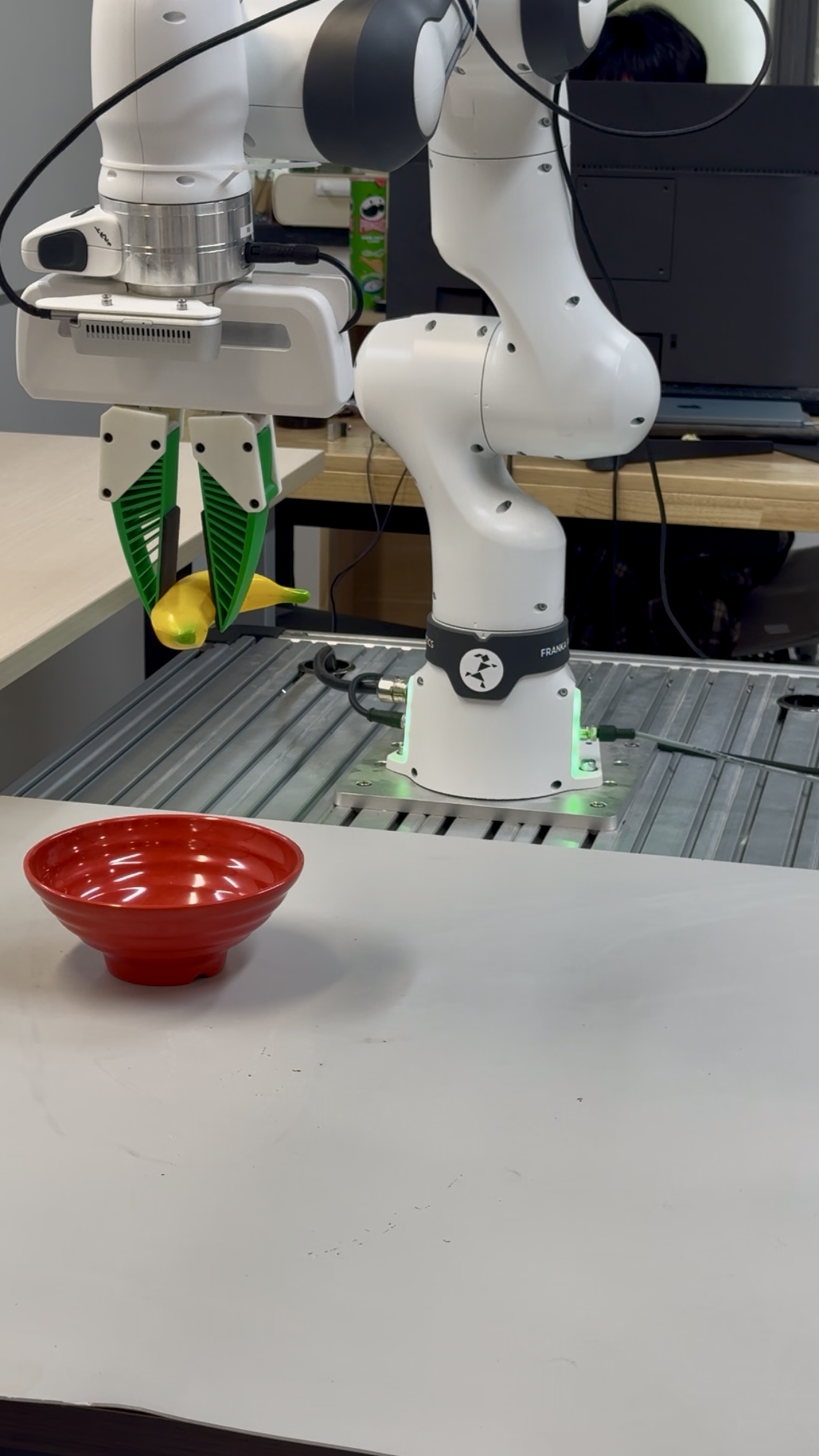}
    \end{subfigure}\hfill
    \begin{subfigure}{0.24\textwidth}
        \includegraphics[width=\linewidth, trim=0 18cm 0 18cm, clip]{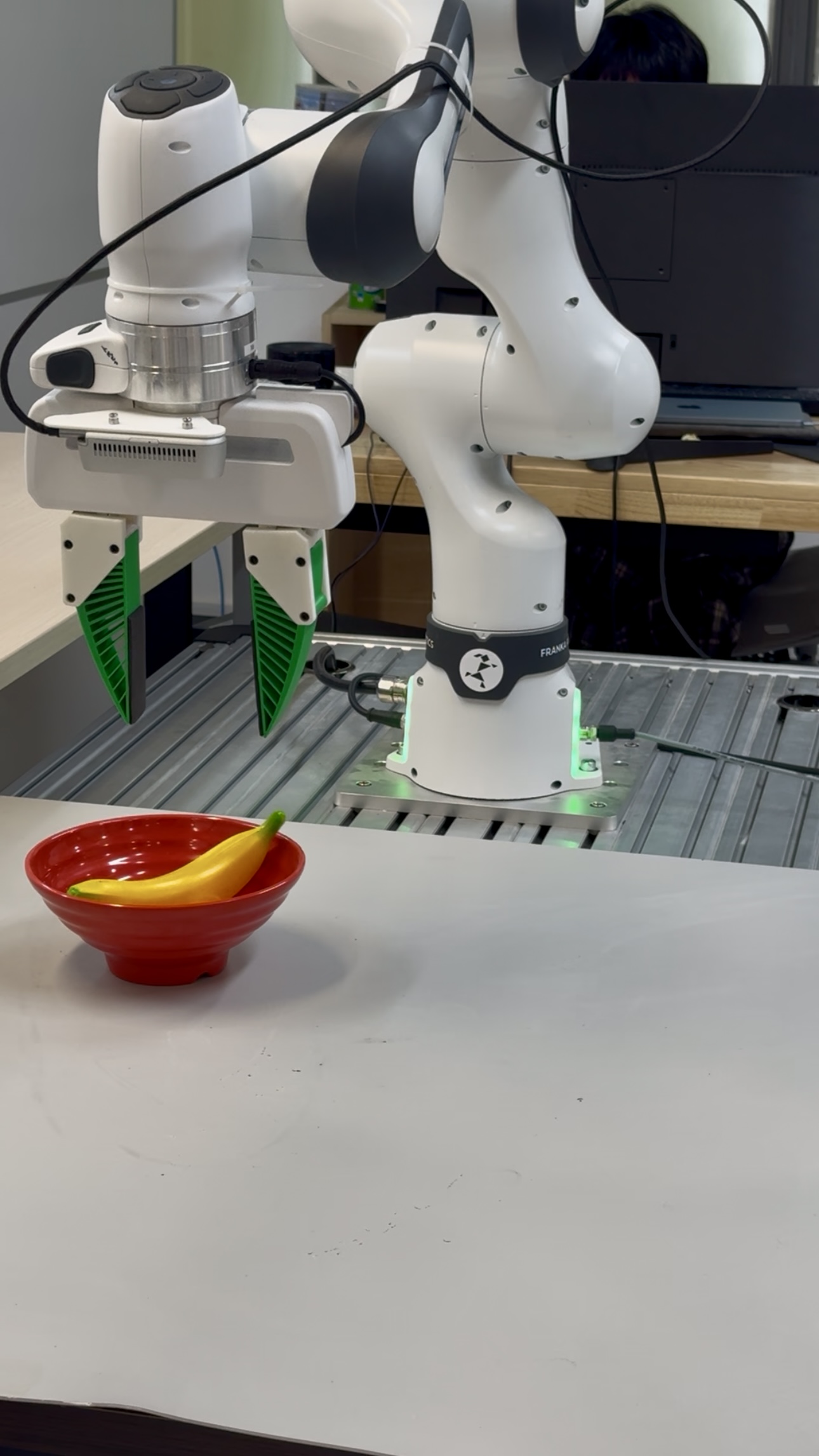}
    \end{subfigure}
    
    \vspace{0.1em}
    
    \begin{subfigure}{0.24\textwidth}
        \includegraphics[width=\linewidth, trim=0 18cm 0 18cm, clip]{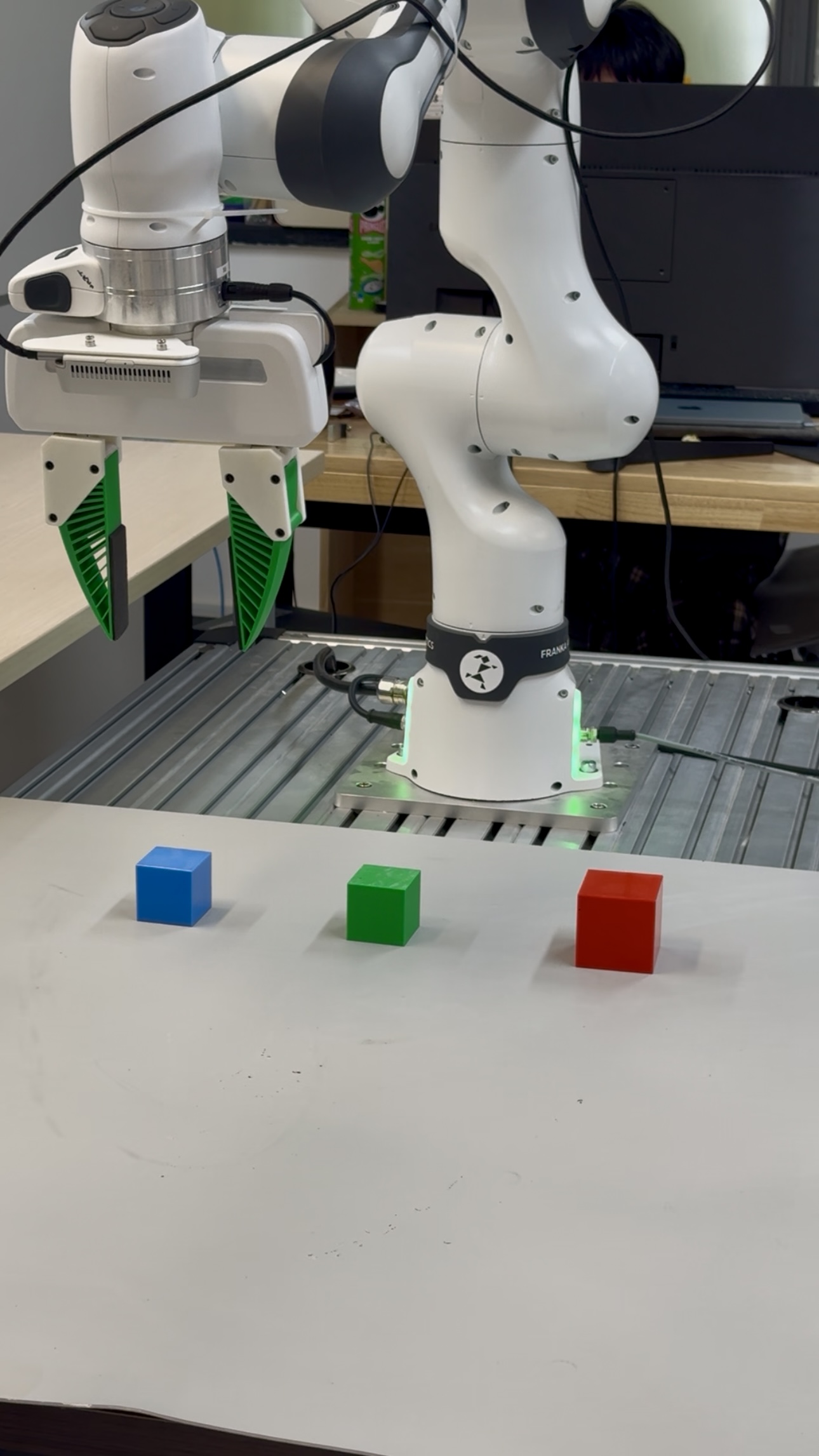}
    \end{subfigure}\hfill
    \begin{subfigure}{0.24\textwidth}
        \includegraphics[width=\linewidth, trim=0 18cm 0 18cm, clip]{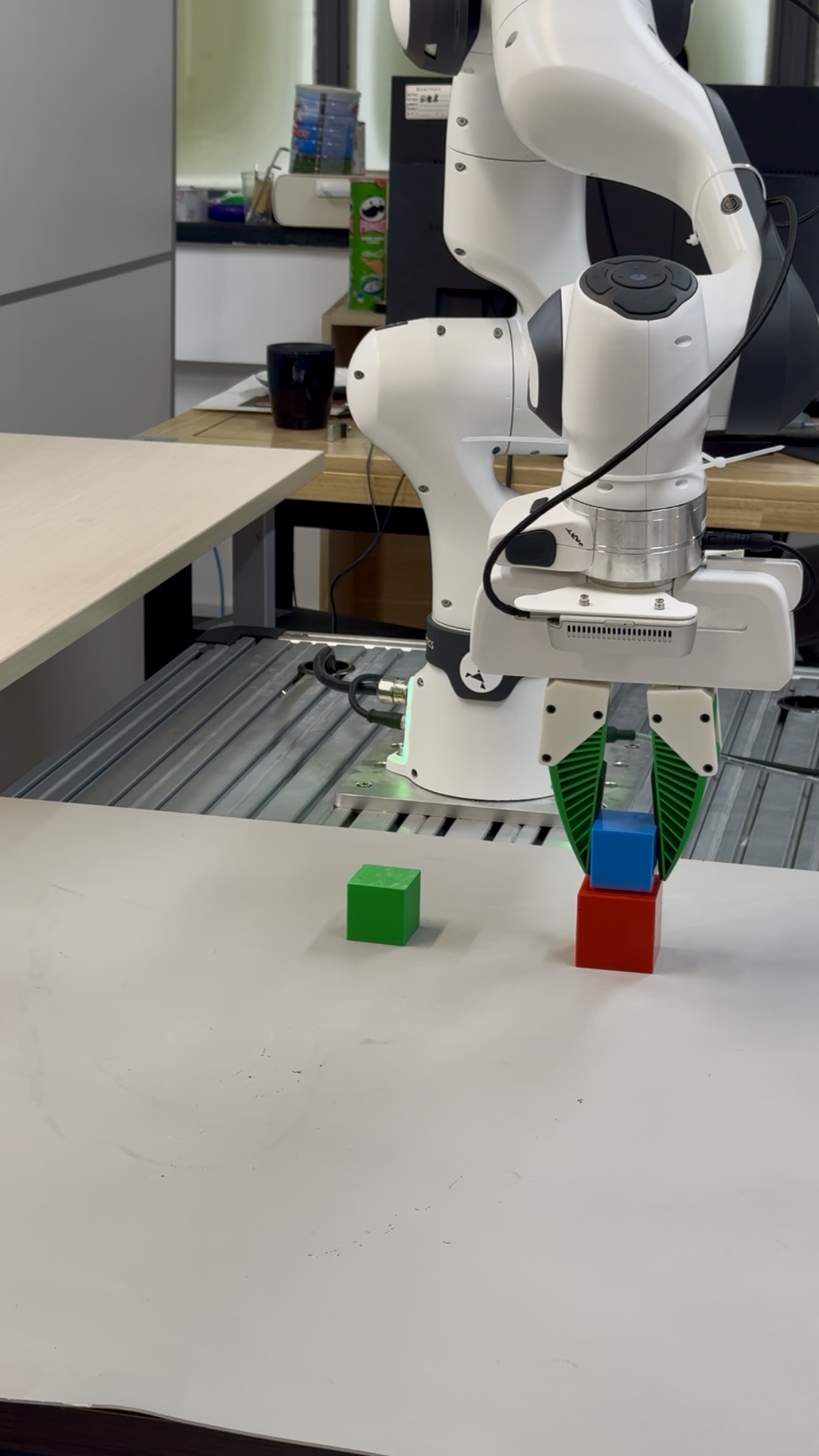}
    \end{subfigure}\hfill
    \begin{subfigure}{0.24\textwidth}
        \includegraphics[width=\linewidth, trim=0 18cm 0 18cm, clip]{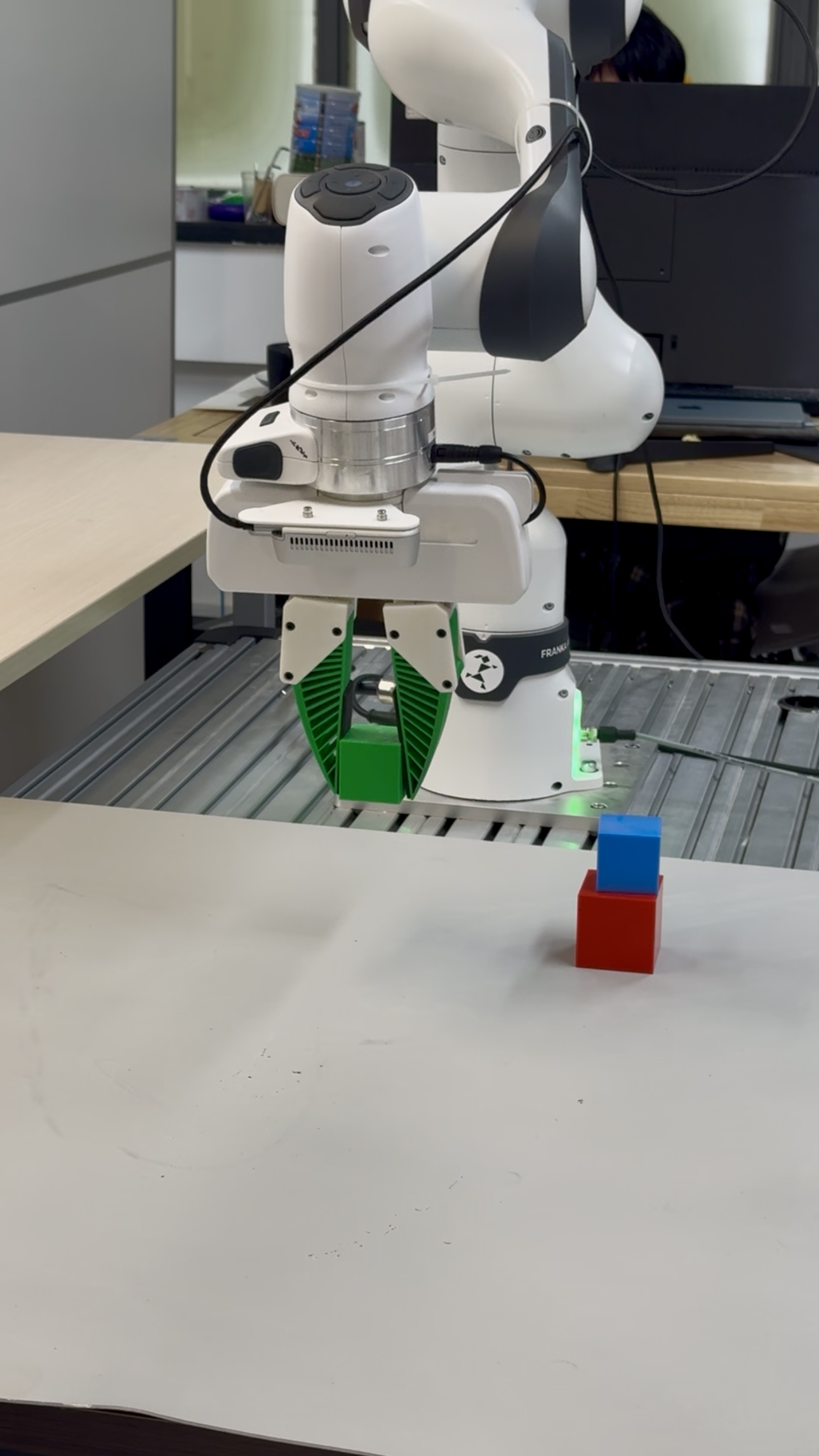}
    \end{subfigure}\hfill
    \begin{subfigure}{0.24\textwidth}
        \includegraphics[width=\linewidth, trim=0 18cm 0 18cm, clip]{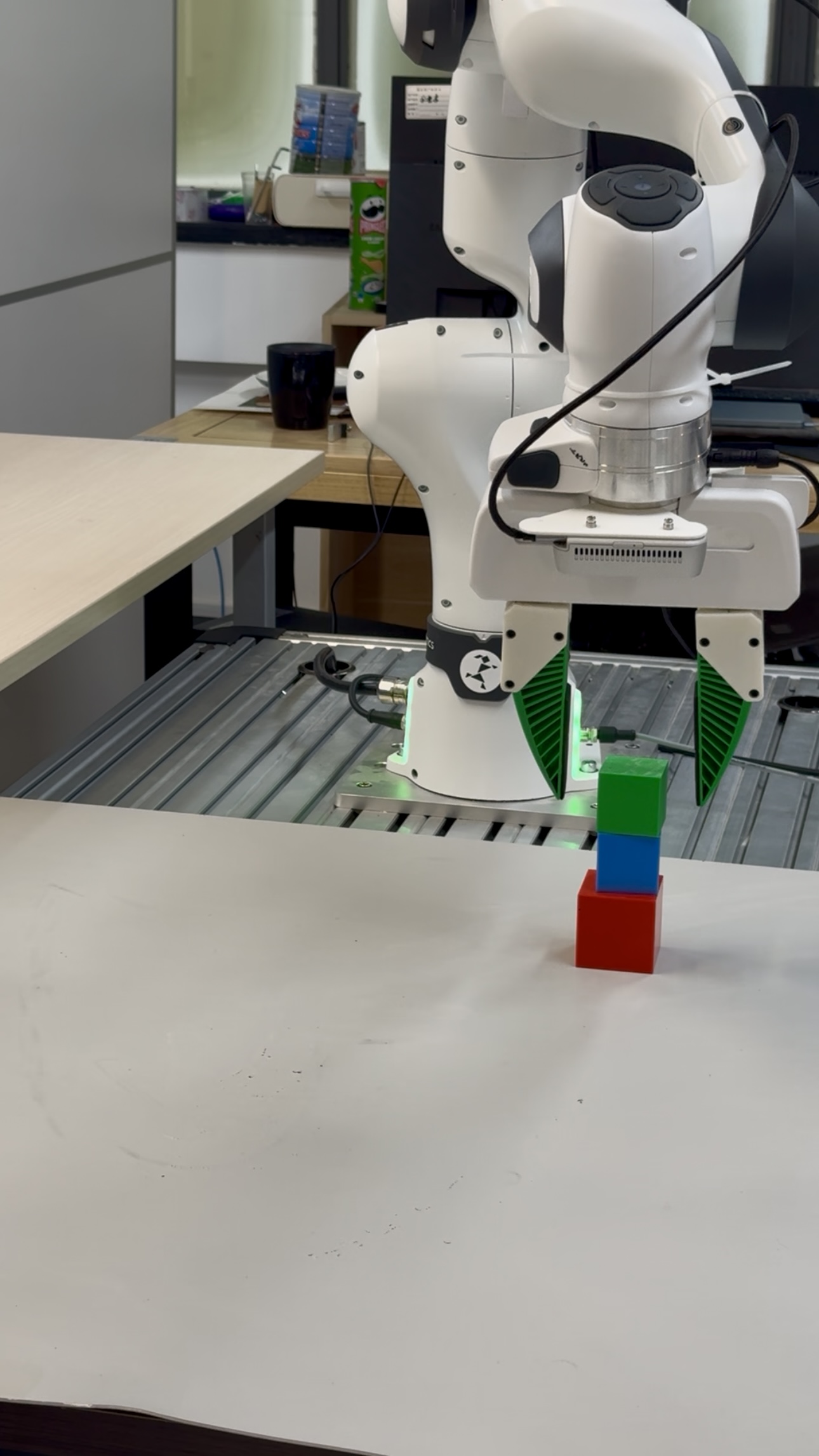}
    \end{subfigure}
    
    \vspace{0.1em}
    
    \begin{subfigure}{0.24\textwidth}
        \includegraphics[width=\linewidth, trim=0 18cm 0 18cm, clip]{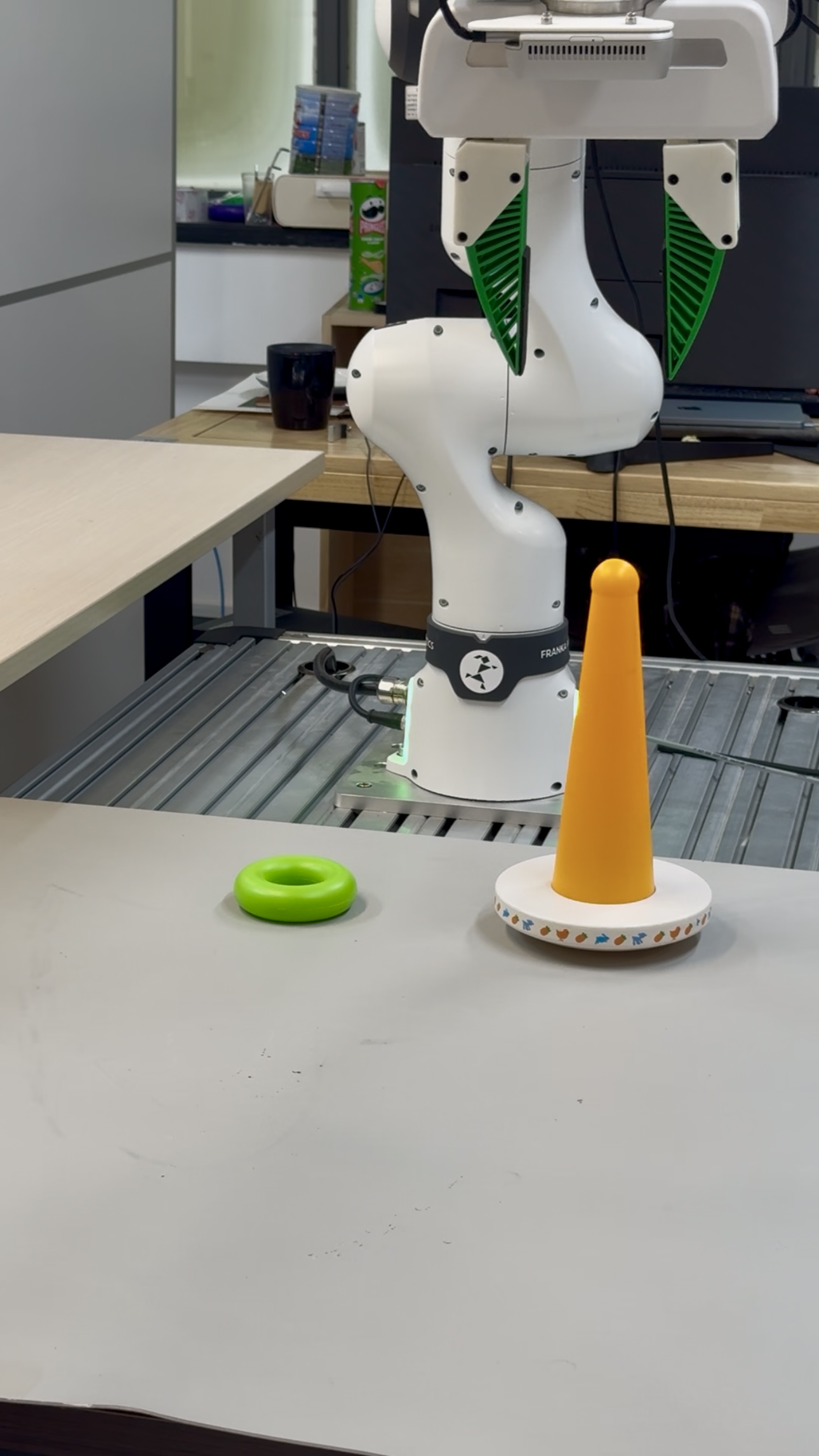}
    \end{subfigure}\hfill
    \begin{subfigure}{0.24\textwidth}
        \includegraphics[width=\linewidth, trim=0 18cm 0 18cm, clip]{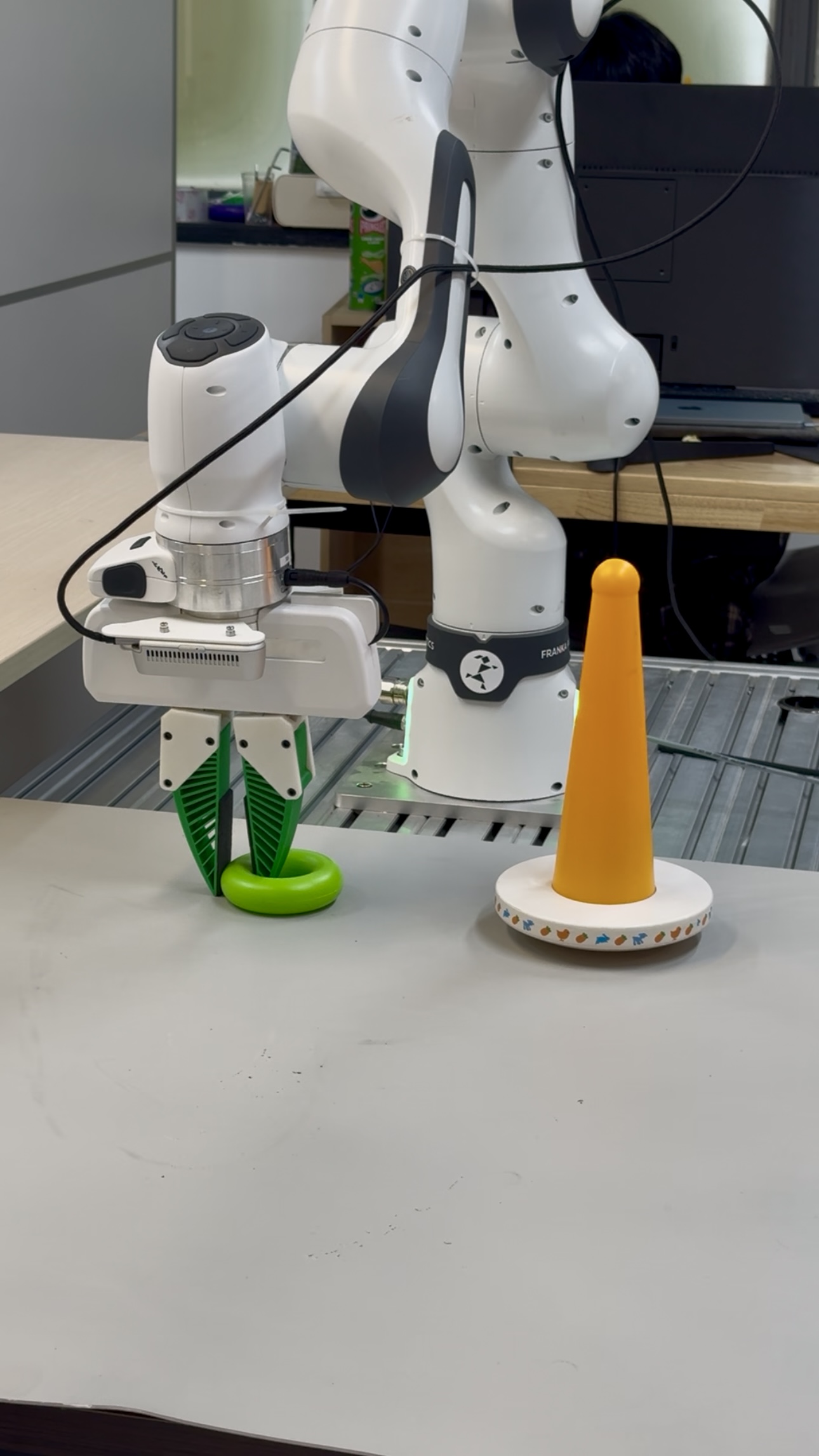}
    \end{subfigure}\hfill
    \begin{subfigure}{0.24\textwidth}
        \includegraphics[width=\linewidth, trim=0 18cm 0 18cm, clip]{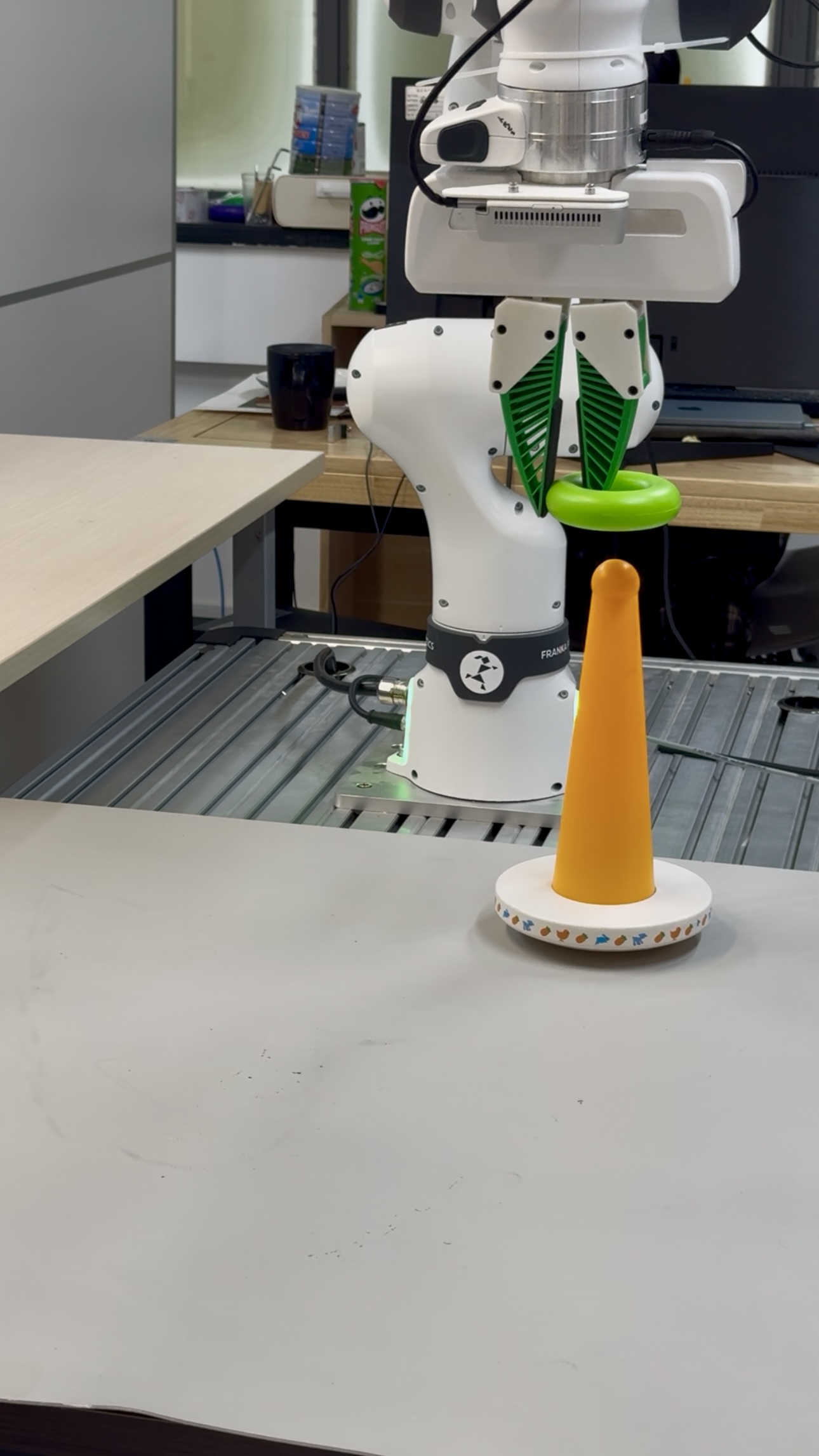}
    \end{subfigure}\hfill
    \begin{subfigure}{0.24\textwidth}
        \includegraphics[width=\linewidth, trim=0 18cm 0 18cm, clip]{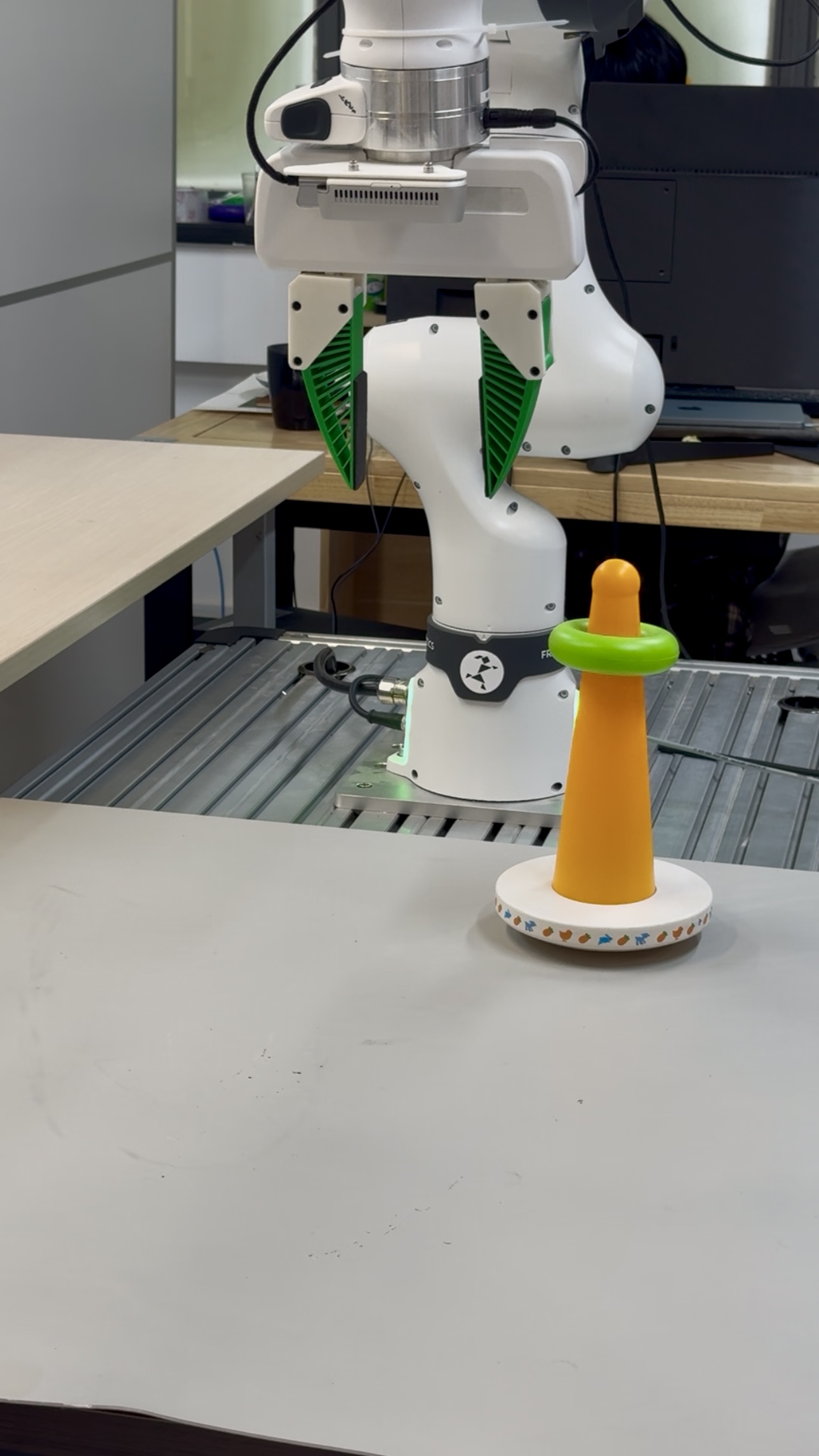}
    \end{subfigure}
    
    \caption{Real-world execution sequences of \textit{ECHO} on three tabletop manipulation tasks: pick-and-place, block stacking, and ring insertion. Each row shows one complete rollout on the Franka platform.}
    \label{fig:all_photos_appendix}
\end{figure}

\textbf{Results and Qualitative Analysis.}
As shown in Table~\ref{tab:real_world_results}, \textit{ECHO} consistently outperforms Vanilla $\pi_0$ across all three tasks. Figure~\ref{fig:robot_photo} illustrates the physical environment setup, and Figure~\ref{fig:all_photos_appendix} provides qualitative execution sequences of \textit{ECHO} completing the long-horizon manipulation steps.

\section{Limitations}
\label{app:limitations}
While \textit{ECHO} demonstrates consistent gains in long-horizon manipulation, its performance still depends on the quality of semantic keyframe extraction and the relevance of retrieved memories. In practice, this is mitigated by the VLM-guided consistency check and similarity-modulated residual injection, which reduce the influence of mismatched or noisy memory entries. Our real-world evaluation focuses on representative tabletop manipulation tasks rather than fully open-ended household scenarios. Scaling \textit{ECHO} to larger lifelong memory banks and more diverse physical environments remains an important direction for future work.

\clearpage



\end{document}